\documentclass{article}
\usepackage{paperstyle,times}

\usepackage{amsmath,amsfonts,bm}

\def\eqref#1{equation~\ref{#1}}

\def\1{\bm{1}}

\DeclareMathAlphabet{\mathsfit}{\encodingdefault}{\sfdefault}{m}{sl}
\SetMathAlphabet{\mathsfit}{bold}{\encodingdefault}{\sfdefault}{bx}{n}

\usepackage{hyperref,url,amsmath,amssymb,booktabs,multirow,graphicx,microtype}
\usepackage[font=small,labelfont=bf]{caption}
\usepackage[font=small]{subcaption}
\newcommand{\panel}[1]{#1.pdf}
\usepackage{wrapfig,array,tabularx,needspace,placeins}
\newcolumntype{Y}{>{\raggedright\arraybackslash}X}

\title{Spectral Steering: Local Inverse Geometry for Frozen Diffusion Generators}
\author{
Angelo Moroncelli$^{1,2}$ \quad
Alessandro Leanza$^{1,2}$ \quad
Loris Roveda$^{1,3}$\\[6pt]
$^1$ University of Applied Sciences and Arts of Southern Switzerland (SUPSI), IDSIA\\
$^2$ Universit\`a della Svizzera italiana (USI)\\
$^3$ Politecnico di Milano
}

\newcommand{\Jh}{J_{h,t}}
\newcommand{\Frem}{F_{t\rightarrow 0}}
\newcommand{\dx}{\Delta x}
\newcommand{\gpred}{G_{\mathrm{cos}}^{\mathrm{pred}}}

\begin{document}
\maketitle
\begin{abstract}
A frozen diffusion generator may need to change several output properties while preserving others. We study when guidance faithfully realizes such a structured request and whether local inverse correction improves its direction. Our account links this benefit to the interaction between the request and the objective-specific remaining-sampler Jacobian: conditioning alone is insufficient. Spectral steering combines a request-specific cosine-gain predictor with damped inverse correction and fixed-time relinearization for finite edits. Scalar nulls and matched-movement controls test the predicted mechanism. Across 8,923 instances in a retrospectively identified local regime, predicted and realized gains correlate at Spearman $0.79$, versus $-0.11$ for condition number. Experiments span images, audio, robotics, and control and planning, including complex character motion generation. On 50 paired inpainting cases, four corrections improve unobserved-pixel PSNR over Adam by 3.25 dB at equal total derivative traversals. Together, the analysis and controls connect request-dependent geometry to directional gains and establish where these gains translate into downstream performance.
\end{abstract}

\section{Introduction}
\begin{wrapfigure}[15]{r}{0.43\textwidth}
    \centering
    \includegraphics[width=\linewidth]{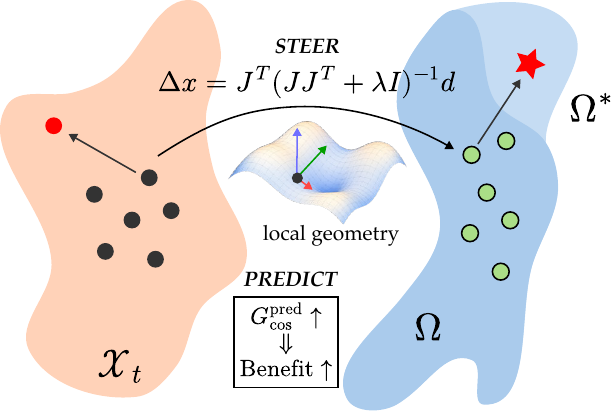}
    \caption{\textbf{Spectral steering of a frozen generator.}
    Spectral steering uses a damped local inverse
    of the requested downstream properties.}
    \label{fig:families}
\end{wrapfigure}
Frozen diffusion generators must often satisfy new tasks without retraining (Fig.~\ref{fig:families}). Differentiable guidance propagates output objectives to sampling states~\citep{bansal2023universal,chung2023diffusion,yu2023freedom}, while differentiation through generation enables direct latent optimization~\citep{wallace2023end,benhamu2024dflow,karunratanakul2024optimizing}. We ask when gradients faithfully realize a \emph{specified pattern of downstream changes}, and whether directional correction gains can be predicted.

Requests combine change and preservation: brighten one region while holding another fixed, alter frequency bands while preserving context, or move a character's wrist while keeping its feet planted. Loss decrease alone does not describe which components move. We evaluate \emph{request fidelity} alongside task success.

\begin{figure}[t]
\centering
\includegraphics[width=\linewidth]{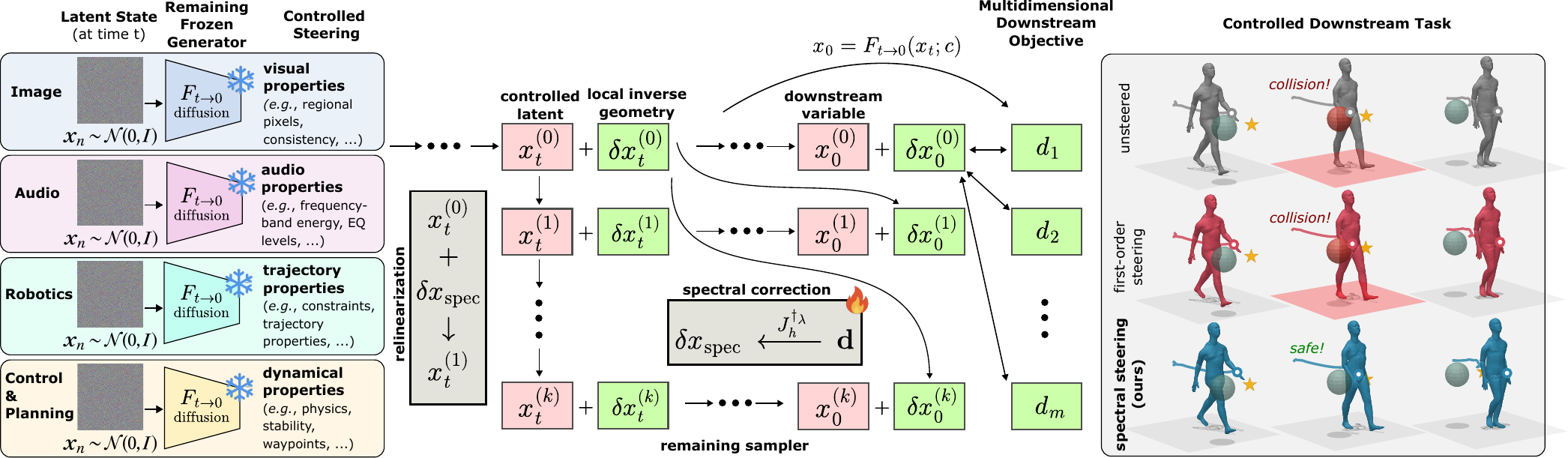}
\caption{\textbf{From gradient pullback to local inversion.} The frozen remaining sampler maps an intervention to downstream properties. FO reweights a request by $J_hJ_h^\top$; the damped inverse compensates unequal gains. K2/K4 refresh the Jacobian and residual at the same diffusion time. Right: unsteered, matched FO, and K4 character motions at two common instants; red spheres mark skeleton--obstacle intersection.}
\label{fig:method}
\end{figure}

The relevant local input--output map is the Jacobian $J_h$ of the selected properties through the entire remaining sampler. Pulling a request $d$ back as $\eta J_h^\top d$ predicts $\eta J_hJ_h^\top d$, so unequal singular gains can alter its relative components. Spectral steering uses classical damped least squares~\citep{levenberg1944method,marquardt1963algorithm} to approximately compensate these gains. For finite edits, it refreshes the Jacobian and residual toward the original target at fixed diffusion time (Fig.~\ref{fig:method}).

Correction gains depend on the request. Even an ill-conditioned Jacobian preserves a supported single-mode request's direction. Distortion depends jointly on the spectrum and the request's modal composition. We express this dependence as a prediction of cosine improvement. An exact scalar null and matched-movement controls make the account experimentally testable beyond aggregate method rankings.

Downstream systems require selected quantities to change by specified amounts while others remain fixed. Character motion requires coordinated joint changes and preservation; robotics adds execution dynamics, and content editing requires preserved context. Our experiments connect predicted directional gains to realized edits and task outcomes, including failures of local approximation or downstream transfer.

Our contributions are:

\noindent\textbf{(1) Request-specific geometry of faithful steering.}
We characterize how the objective-specific remaining-sampler geometry distorts structured requests, identifying the role of modal composition and an exact scalar null after magnitude matching (Sections~\ref{sec:method} and~\ref{sec:mech}).

\noindent\textbf{(2) Predicting directional gains and refining finite edits.}
Our score $\gpred$ predicts local cosine improvement over first-order guidance. Spectral correction implements the damped inverse, and fixed-time relinearization refreshes the geometry for finite requests (Sections~\ref{sec:method} and~\ref{sec:predictor}).

\noindent\textbf{(3) Controlled evidence across generative domains.}
Movement controls and optimizer comparisons test directional gains, task success, and executed outcomes across images, audio, robotics, and control and planning---beginning with complex character motion generation (Section~\ref{sec:results}).

\section{Related Work}\label{sec:related_work}
\paragraph{Gradient guidance.}
Classifier guidance, Universal Guidance, Loss-Guided Diffusion and DPS condition sampling through gradients~\citep{dhariwal2021diffusion,bansal2023universal,song2023loss,chung2023diffusion}; FreeDoM uses energy-based guidance~\citep{yu2023freedom}. These methods establish general differentiable conditioning; our question is how faithfully a gradient transmits a specified vector displacement.

\paragraph{End-to-end optimization.}
Earlier text-guided face editing uses GAN inversion~\citep{xia2021tedigan}. Full-process differentiation is established: FlowGrad, DOODL and D-Flow optimize generative states through the resulting output~\citep{liu2023flowgrad,wallace2023end,benhamu2024dflow}; DNO applies this principle to motion and DITTO to music~\citep{karunratanakul2024optimizing,novack2024ditto}. D-Flow also analyzes the generator Jacobian's action on output gradients. OC-Flow formulates optimal control~\citep{wang2025ocflow}, and posterior and end-to-end guidance have been connected theoretically~\citep{blasingame2025greed}. We examine how a prescribed vector request interacts with this geometry, predicting directional correction gains and testing them under movement controls.

\paragraph{Geometry and inverse guidance.}
DDRM and DDNM exploit measurement structure~\citep{kawar2022denoising,wang2023zero}; $\Pi$GDM uses measurement-model pseudoinverses for conditional-score estimation~\citep{song2023pseudoinverse}. Spherical Gaussian and manifold-preserving guidance constrain intermediate or data-manifold geometry~\citep{yang2024guidance,he2024manifold}; Park et al.\ analyze latent singular directions~\citep{park2023understanding}. JAPS studies denoiser Jacobians, while CLAMP applies denoiser-pulled curvature and manifold-aligned damping to posterior correction~\citep{hen2026jacobian,shin2026geometry}. We study the selected downstream representation composed with the complete remaining sampler, asking which requests benefit from inverse correction. The contribution combines a request-dependent prediction with scalar nulls, movement controls, and finite-edit evaluation. The damped least-squares update supplies the correction; it does not define a posterior sampler.

\paragraph{Planning and policy steering.}
Offline demonstration learning highlights the role of data and evaluation in manipulation~\citep{robomimic2021}. Diffuser and Diffusion Policy generate trajectories and actions~\citep{janner2022planning,chi2023diffusionpolicy}; SafeDiffuser and JM2D address constrained planning~\citep{xiao2025safediffuser,jung2025joint}. DynaGuide uses a dynamics model to steer policies, whereas DSRL learns a latent-noise controller from environment interaction~\citep{du2025dynaguide,wagenmaker2025steering}. Spectral steering requires a differentiable specification without training another controller. Execution and safety remain separate evaluation requirements. Appendix~\ref{app:main-details} summarizes the control interpretation.

\section{Objective-Specific Local Inverse Geometry}\label{sec:method}
\subsection{Structured requests and local geometry}
At fixed diffusion time $t$, let $x_t\in\mathbb R^D$ be the intervention state and $z(x_t)=h(\Frem(x_t;c))\in\mathbb R^Z$ the downstream properties. The map includes the complete remaining sampler, decoding, and task-specific coordinate scaling; conditioning and remaining randomness are fixed. Its Jacobian is
\begin{equation}
\Jh=\frac{\partial(h\circ\Frem)(x_t;c)}{\partial x_t}.
\label{eq:full-jacobian}
\end{equation}
For target $z_\star=z(x_t)+d$, the request $d=[d_{\rm change}^\top,0_{\rm preserve}^\top]^\top$ specifies changes and held properties. We seek a finite displacement $z(x_t+\dx)-z(x_t)\approx d$, using the local model
\begin{equation}
z(x_t+\dx)-z(x_t)\approx\Jh\dx.
\label{eq:linearization}
\end{equation}
Preservation is a soft least-squares target; omitted properties are uncontrolled. Gradient descent on $\frac12\|z(x)-z_\star\|_2^2$ gives
\begin{equation}
\dx_{\rm FO}=\eta\Jh^\top d,\qquad
\Delta z_{\rm FO}\approx\eta\Jh\Jh^\top d,\quad\eta>0.
\label{eq:fo}
\end{equation}
This is a descent direction when nonzero, but its predicted response need not align with $d$. A scalar step size changes the response magnitude without correcting the relative components. The relevant geometry depends on both the generator and the downstream representation: changing the measurements or their scaling changes the operator being inverted.

\subsection{Damped inversion and request-specific prediction}
For $\lambda>0$, spectral steering solves the regularized local problem
\begin{align}
\dx_{\rm spec}&=\arg\min_\delta\;\tfrac12\|\Jh\delta-d\|_2^2+\tfrac\lambda2\|\delta\|_2^2,\label{eq:local-ls}\\
&=(\Jh^\top\Jh+\lambda I_D)^{-1}\Jh^\top d=\Jh^\top(\Jh\Jh^\top+\lambda I_Z)^{-1}d.\label{eq:woodbury}
\end{align}
One correction is \textbf{K1}. The equivalent primal and dual forms solve in state and property space, respectively. For compact SVD $\Jh=U\Sigma V^\top$ and request $d=\sum_i a_i u_i+d_\perp$, with $U^\top d_\perp=0$, the predicted responses are
\begin{equation}
\Jh\dx_{\rm FO}=\eta\sum_i\sigma_i^2a_i u_i,\qquad
\Jh\dx_{\rm spec}=\sum_i\frac{\sigma_i^2}{\sigma_i^2+\lambda}a_i u_i.
\label{eq:svd-response}
\end{equation}
For $\sigma_i^2\gg\lambda$, inverse gains approach one; weak modes remain attenuated and $d_\perp$ is locally unreachable. A supported single-mode request has collinear responses regardless of the condition number. Directional benefit therefore depends on which modes the request occupies, not just the spectrum. For $Z=1$, magnitude matching makes the latent updates identical (Appendix~\ref{sec:scalar-null}).

Writing $G_h=\Jh\Jh^\top$, we predict the directional gain by
\begin{equation}
\gpred=\cos\!\left(G_h(G_h+\lambda I)^{-1}d,d\right)-\cos(G_hd,d).
\label{eq:predictor}
\end{equation}
For nonzero requests and responses, the score predicts local cosine gain before editing. Different requests can have different predicted gains at the same Jacobian. Agreement with nonlinear edits is tested empirically; magnitude and task utility are evaluated separately.

\subsection{Fixed-time relinearization and computation}
\begin{figure}[t]
    \centering
    \begin{subfigure}[t]{0.45\linewidth}
        \centering
        \includegraphics[width=\linewidth]
        {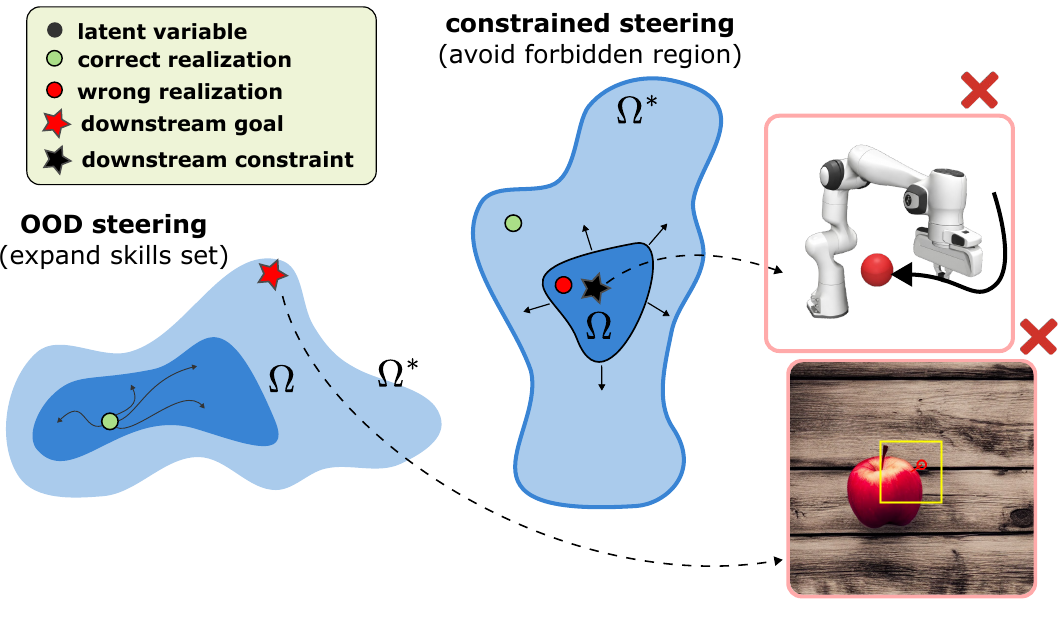}
        \caption{}
        \label{fig:relinearization-a}
    \end{subfigure}
    \hfill
    \begin{subfigure}[t]{0.53\linewidth}
        \centering
        \includegraphics[width=\linewidth]
        {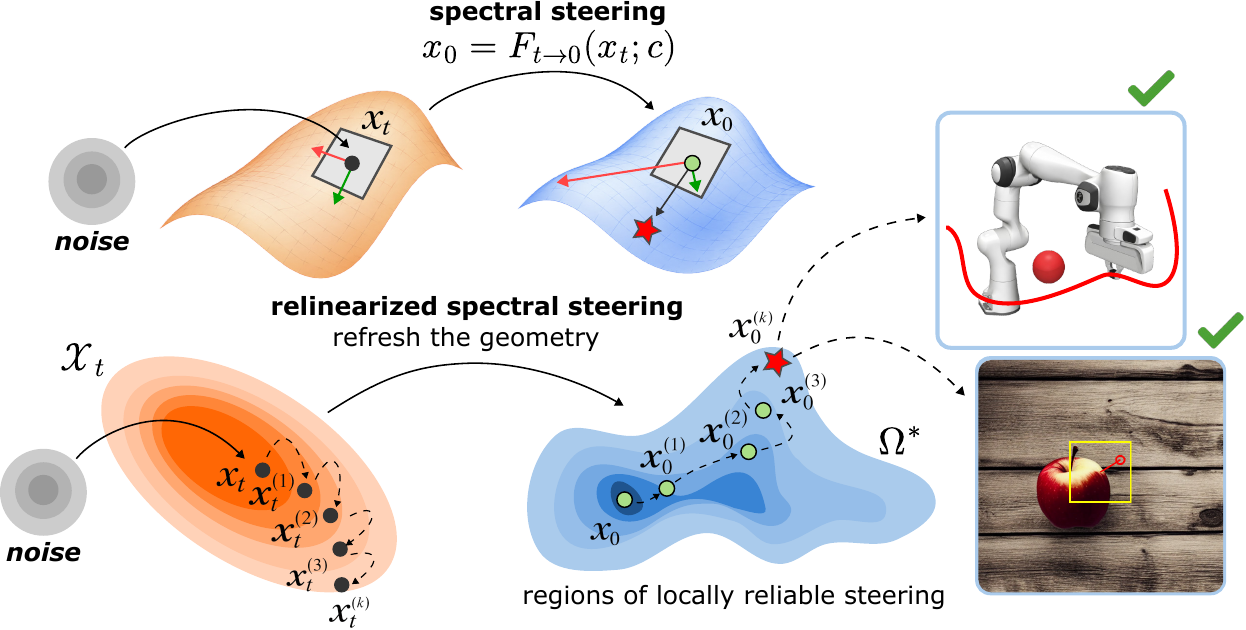}
        \caption{}
        \label{fig:relinearization-b}
    \end{subfigure}
    \caption{\textbf{Structured steering and fixed-time relinearization.} (a) Changes and preservation must be realized jointly through the frozen generator. (b) Successive corrections refresh the Jacobian and residual toward the original target at the same diffusion time. $\Omega^\star$ denotes the target action set that becomes reachable after steering.}
    \label{fig:relinearization-detail}
\end{figure}

K2/K4 repeat the correction toward the original target. At $x_t^{(k)}$, recompute $J_k=\partial z/\partial x_t^{(k)}$, set $r_k=z_\star-z(x_t^{(k)})$, and update
\begin{equation}
 x_t^{(k+1)}=x_t^{(k)}+J_k^\top(J_kJ_k^\top+\lambda_k I)^{-1}r_k,
 \qquad\lambda_k=0.01\sigma_1(J_k)^2.
\label{eq:relinearize}
\end{equation}
The index $k$ counts corrections, not denoising steps; the final output is $\Frem(x_t^{(K)};c)$. Each candidate completes the same remaining sampler, measures its residual relative to the original target, and updates the intervention state. Keeping that target fixed avoids requesting the full edit again at every correction. Refreshing the local model addresses nonlinear error without guaranteeing convergence (Fig.~\ref{fig:relinearization-detail}). The coefficient $0.01$ is fixed; ``adaptive'' damping means refreshing $\sigma_1^2$ at each correction, rather than a line search or a latent-norm bound.

Explicit solvers use reduced-SVD filtering without hard rank truncation and obtain $\sigma_1$ from the Jacobian. Matrix-free solvers use up to 16 request-started Golub--Kahan iterations, estimate $\sigma_1$ from the projected matrix, and solve the damped problem there. Each iteration uses one VJP and one JVP through the remaining sampler: one correction costs up to 16 of each, and $K$ corrections up to $16K$ of each. ``Direct'' inversion refers to the local inverse relation, not iteration-free computation. Numerical-support thresholds are diagnostic, separate from damping; setting-specific thresholds, stopping rules, and solver variants are in Appendix~\ref{app:solver-implementation}.

\section{Experimental Protocol}\label{sec:protocol}
Vision uses a CelebA-HQ DDPM~\citep{ho2020denoising} for restoration and Stable Diffusion 1.5~\citep{rombach2022high} for structured edits. Audio covers MAESTRO waveform excerpts~\citep{hawthorne2018enabling} and generated mel spectrograms; robotics measures MuJoCo execution~\citep{6386109}. Control and planning begins with 60-frame MDM character motions~\citep{tevet2023human}, followed by trajectory and dynamical-system tasks. Table~\ref{tab:cross-domain} identifies models; Appendix~\ref{sup:sec:task-objectives} defines objectives and endpoints.

\paragraph{Matching and baselines.}
FO and K1 share the initial state, time, sampler, objective, and request. FO matches K1's \emph{predicted objective-space} movement with $\eta=\|\Jh\dx_{\rm K1}\|_2/\|\Jh\Jh^\top d\|_2$ for nonzero denominator, using the task's scaled coordinates. It reuses K1's Jacobian or corresponding Krylov quantities; this isolates direction, not standalone cost. A separate control matches latent displacement. SGD, Adam, L-BFGS and domain-specific baselines use fixed, pilot-selected, or evaluation-selected settings as documented in Appendix~\ref{app:vision-populations}; the main table is not a common-budget ranking.

\paragraph{Evaluation.}
We distinguish request fidelity, application outcomes, and executed behavior. Paired comparisons are used where available. Compute comparisons separately count vector--Jacobian products (VJPs), total VJP/Jacobian--vector-product (JVP) traversals, or network evaluations; these do not establish equal wall-clock cost. Shared-task adaptations are distinguished from native baseline protocols. A matrix-free spectral solve uses up to 16 VJPs and 16 JVPs, whereas each SGD/Adam step uses one VJP; explicit implementations have different derivative representations. Sample sets, uncertainty, solver mappings, and qualitative selection rules are specified with the results and in the appendix.

\section{Results and Evaluation}\label{sec:results}
\subsection{Correcting modal gains improves structured responses}\label{sec:mech}
\begin{figure}[t]
\centering
\makebox[\textwidth][c]{%
\begin{subfigure}[t]{0.32\textwidth}
\centering\includegraphics[width=\linewidth]{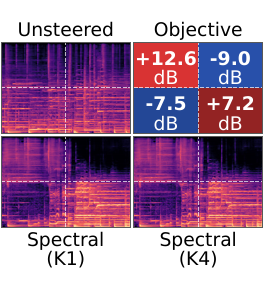}
\caption{Audio request}\label{fig:audio-example}
\end{subfigure}%
\hspace{.00666\textwidth}%
\begin{subfigure}[t]{0.22\textwidth}
\centering\includegraphics[width=\linewidth]{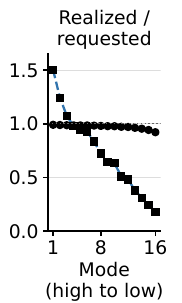}
\caption{Modal correction}\label{fig:mech:modes}
\end{subfigure}%
\hspace{.00666\textwidth}%
\begin{subfigure}[t]{0.22\textwidth}
\centering\includegraphics[width=\linewidth]{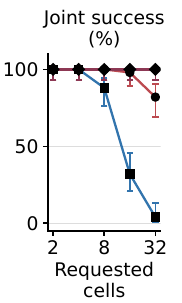}
\caption{Joint success}\label{fig:mech:complexity}
\end{subfigure}%
\hspace{.00666\textwidth}%
\begin{subfigure}[t]{0.22\textwidth}
\centering\includegraphics[width=\linewidth]{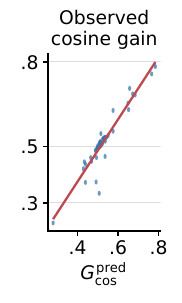}
\caption{Motion predictor}\label{fig:mech:predictor}
\end{subfigure}%
}
\par\vspace{2pt}\includegraphics[width=\linewidth]{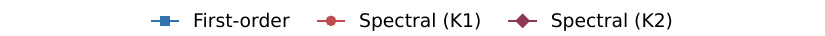}
\caption{\textbf{Testing the directional mechanism.} (a) Four-region audio request; spectrograms share a color scale (time right, frequency up). (b) FO overweights sensitive modes; inverse responses lie within 5\% of each displayed request. (c) At 32 audio cells, FO/K1/K2 attain 4/82/100\% success ($n=50$; 95\% intervals). (d) Predicted versus realized cosine gain on 50 character-avoidance seeds ($r=0.94$, $\rho=0.92$); red line: linear fit. Appendix~\ref{app:predictor-audit} reports the cross-domain analysis.}
\label{fig:mech}
\end{figure}

Figure~\ref{fig:mech}(a,b) connects a four-region audio request to the modal mechanism: the pullback overweights sensitive modes, while inverse responses lie within 5\% of the request in the displayed example. The scalar null supplies a falsifiable control: all 935 tested pairs tie to machine precision at $Z=1$. Benefits emerge with multiple coordinates. At 32 audio cells, joint success is 4\% for FO, 82\% for K1, and 100\% for K2 ($n=50$), showing improved simultaneous realization and the value of relinearization.
\subsection{Local geometry predicts directional gains}\label{sec:predictor}
On 50 character-motion avoidance seeds, predicted and observed cosine gains correlate at $r=0.94$, $\rho=0.92$ (Fig.~\ref{fig:mech}(d)). Across 8,923 instances in 27 vision, audio, and robotics populations within the measured local regime, $r=0.75$ and $\rho=0.79$, versus $-0.04$ and $-0.11$ for log condition number. Centering within populations retains $\rho=0.77$, and Pearson association is positive in 26 of 27 populations, supporting an association beyond pooled between-task differences (Appendix~\ref{app:predictor-audit}). The score is computed before editing, but regime selection is retrospective; this supports a local diagnostic, not a validated acceptance rule.

\paragraph{Matched movement.}\label{sec:direction}
Across 576 instances with latent norms agreeing within $4\times10^{-14}$, K1 improves request realization on 100\% of inpainting images, 85\% of audio excerpts, and 93\% of Panda instances. Interpolating toward the inverse at fixed predicted response norm improves reconstruction by 5.34 dB, with improvements on all 50 images at every tested step (Appendix~\ref{app:direction}). These controls isolate a directional benefit. They also separate the proposed mechanism from simply taking a larger step: the latent-norm control fixes intervention size, while the interpolation fixes predicted downstream movement (Appendix~\ref{app:direction}).

\subsection{Vision: reconstruction and structured edits}\label{sec:image}
\begin{table*}[t]
\centering\small
\caption{\textbf{General optimizers across domains.} Bold: best displayed value, including rounded ties. Settings and selected budgets vary by task. ns: no statistically significant difference between the two marked methods in the same row. GD/Adam learning rates selected by a hyperparameter search, L-BFGS learning rate selected on the evaluation set. Character optimizers match K4's nominal backward-pass count. Protocols and checkpoints: Appendices~\ref{app:vision-populations}, \ref{app:motion}, and~\ref{app:diffusion-models}.}
\label{tab:cross-domain}
\setlength{\tabcolsep}{2pt}
\renewcommand{\arraystretch}{1.06}
\begin{tabular}{@{}l>{\raggedright\arraybackslash}p{1.15in}rrrrrrl@{}}
\toprule
\shortstack[l]{Data / simulator\\Diffusion model} & Task / endpoint & Unst. & FO & SGD & Adam & L-BFGS & \shortstack{Spectral\\ (K1)} & \shortstack{Spectral\\ (best K)} \\
\midrule
\multicolumn{9}{@{}l}{\textbf{Vision}}\\
\multirow{4}{*}{\shortstack[l]{CelebA-HQ\\DDPM EMA-256}} & Inpainting PSNR $\uparrow$ & 16.55 & 19.61 & 25.57 & 25.98 & 25.76 & 23.69 & \textbf{27.75 (K4)}\\
 & $4\times$ SR PSNR $\uparrow$ & 25.05 & 26.32 & 27.69 & 27.79 & 27.58 & 27.39 & \textbf{27.90 (K2)}\\
 & Deblur PSNR $\uparrow$ & 24.03 & 25.26 & 27.26 & 27.92$^{\mathrm{ns}}$ & 27.90 & 26.61 & \textbf{27.94 (K16)}$^{\mathrm{ns}}$\\
 & \shortstack[l]{Multi-constraint\\PSNR $\uparrow$} & 25.81 & 26.12 & 27.69 & \textbf{28.61} & 28.54 & 26.89 & 27.75 (K4)\\
\cmidrule(lr){1-9}
\shortstack[l]{Generated images\\Stable Diff. 1.5} & \shortstack[l]{Structured edit [0-1] $\uparrow$} & 0.000 & 0.574 & 0.734 & 0.774 & 0.765 & 0.954 & \textbf{1.000 (K8)}\\
\midrule
\multicolumn{9}{@{}l}{\textbf{Audio}}\\
\multirow{2}{*}{\shortstack[l]{MAESTRO 150k}} & Localized EQ err.\ $\downarrow$ & 7.000 & 5.878 & 6.623 & 4.692 & 3.895 & 5.538 & \textbf{2.691 (K16)}\\
 & Bandwidth LSD $\downarrow$ & 12.757 & 13.408 & 13.151 & 26.604 & 13.487 & 12.733 & \textbf{12.730 (K4)}\\
\multirow{2}{*}{\shortstack[l]{Generated mel\\Audio Diffusion}} & EQ-16 success $\uparrow$ & 0.000 & 0.32 & 1.00 & 1.00 & 1.00 & 0.98 & \textbf{1.00 (K2)}\\
 & Selective EQ dB $\downarrow$ & 2.5981 & 1.0883 & 0.0005 & 0.0300 & 0.0005 & 0.3935 & \textbf{0.0004 (K4)}\\
\midrule
\multicolumn{9}{@{}l}{\textbf{Robotics}}\\
\multirow{5}{*}{\shortstack[l]{MuJoCo\\Temporal DDPM}} & Panda reach err.\ cm $\downarrow$ & 3.89 & 2.39 & 6.54 & 2.03 & \textbf{0.63} & 1.06 & 0.74 (K16)\\
 & Panda pose err.\ cm $\downarrow$ & 5.10 & 2.28 & 0.82 & 0.97 & 0.79 & 1.09 & \textbf{0.70 (K4)}\\
 & Cheetah target cm $\downarrow$ & 16.16 & 11.57 & 6.79 & \textbf{4.18} & 5.74 & 6.67 & 4.31 (K16)\\
 & Cheetah traj.\ cm $\downarrow$ & 13.29 & 9.00 & 5.57 & 6.19 & 5.19 & 6.44 & \textbf{4.72 (K16)}\\
 & \shortstack[l]{Panda contact [0-1] $\uparrow$} & 0.000 & 0.769 & 0.766 & 0.851 & \textbf{0.926} & 0.718 & 0.894 (K16)\\
\midrule
\multicolumn{9}{@{}l}{\textbf{Control \& planning}}\\
\multirow{2}{*}{\shortstack[l]{Generated motion\\MDM}} & \shortstack[l]{Char. obstacle [0-1] $\uparrow$} & 0.000 & 0.000 & 0.960$^{\mathrm{ns}}$ & 0.880 & 0.980 & 0.180 & \textbf{0.980 (K16)}$^{\mathrm{ns}}$\\
 & \shortstack[l]{Char. reaching [0-1] $\uparrow$} & 0.000 & 0.000 & 0.480 & 0.000 & 0.960$^{\mathrm{ns}}$ & 0.140 & \textbf{0.960 (K4)}$^{\mathrm{ns}}$\\
\cmidrule(lr){1-9}
\multirow{3}{*}{\shortstack[l]{2D integrator\\Temporal DDPM}} & \shortstack[l]{1-waypoint err. cm $\downarrow$} & 110.86 & 30.35 & 37.69 & 7.48 & 26.34 & 7.41 & \textbf{3.05 (K4)}\\
 & Multi-waypoint [0-1] $\uparrow$ & 0.000 & 0.223 & 0.873& 0.534 & 0.945 & 0.854 & \textbf{1.000 (K4)}\\
 & Point-mass [0-1] $\uparrow$ & 0.000 & 0.983 & 0.999 & 0.997 & 1.000$^{\mathrm{ns}}$ & 0.997 & \textbf{1.000 (K4)}$^{\mathrm{ns}}$\\
\bottomrule
\end{tabular}
\end{table*}

\begin{figure}[t]
\centering
\captionsetup[subfigure]{skip=1pt}
\makebox[\textwidth][c]{%
\begin{subfigure}[t]{.196\textwidth}
\centering\includegraphics[width=\linewidth]{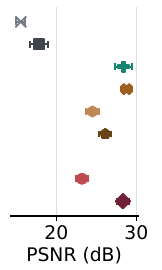}
\caption{Inpainting}\label{fig:vision:inpaint}
\end{subfigure}%
\hspace{.005\textwidth}%
\begin{subfigure}[t]{.196\textwidth}
\centering\includegraphics[width=\linewidth]{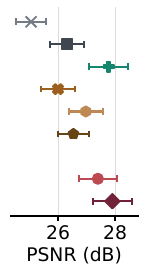}
\caption{$4\times$ SR}\label{fig:vision:sr}
\end{subfigure}%
\hspace{.005\textwidth}%
\begin{subfigure}[t]{.196\textwidth}
\centering\includegraphics[width=\linewidth]{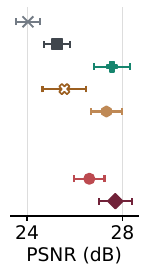}
\caption{Deblurring}\label{fig:vision:deblur}
\end{subfigure}%
\hspace{.005\textwidth}%
\begin{subfigure}[t]{.196\textwidth}
\centering\includegraphics[width=\linewidth]{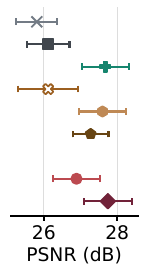}
\caption{Multi-constraint}\label{fig:vision:multi}
\end{subfigure}%
\hspace{.005\textwidth}%
\begin{subfigure}[t]{.196\textwidth}
\centering\includegraphics[width=\linewidth]{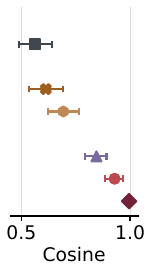}
\caption{Large edit}\label{fig:vision:stress}
\end{subfigure}%
}
\par\vspace{0pt}\includegraphics[width=\linewidth]{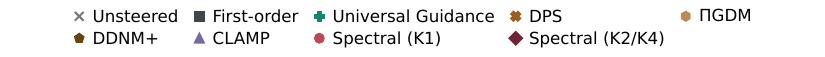}
\par\vspace{2pt}
\begin{subfigure}[t]{\textwidth}
\centering\includegraphics[width=\linewidth]{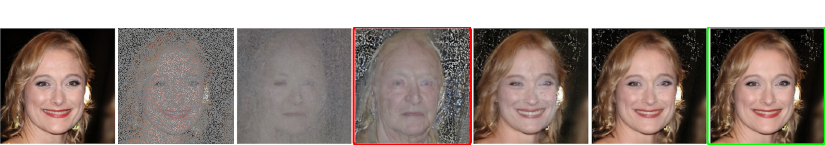}
\caption{A representative inpainting reconstruction}\label{fig:vision:example}
\end{subfigure}
\caption{\textbf{Vision accuracy and reconstruction.} (a--d) CelebA-HQ/DDPM PSNR, 50 reference images per task, 95\% intervals; hollow DPS uses deblur $n=25$ and composite $n=20$. UG uses selected settings; native DPS uses 1000 steps. (e) SD1.5 request cosine at $4\times$ scale (window 10, $n=20$). Deeper spectral rungs are K2 in (b,e), K4 elsewhere. DPS, $\Pi$GDM, and CLAMP in (e) are objective adaptations. (f) One paired reconstruction. Protocols and differing image sets: Appendices~\ref{app:vision-populations} and~\ref{app:ugprotocol}.}
\label{fig:vision-reconciled}
\end{figure}
\paragraph{Generic optimizers.}
In Table~\ref{tab:cross-domain}, K4 reaches 27.75 dB paired inpainting PSNR versus Adam's 25.98. Structured editing gives cosine 0.954 for K1 and 1.000 for K8, versus 0.774 for Adam and 0.765 for L-BFGS. Spectral also leads reported super-resolution PSNR. Adam leads composite restoration; the K16 deblurring comparison is statistically unresolved and costs eight times Adam's network evaluations in that audit.

\begin{figure}[t]
\centering
\makebox[\textwidth][c]{%
\begin{subfigure}[t]{0.245\textwidth}
\centering\includegraphics[width=\linewidth]{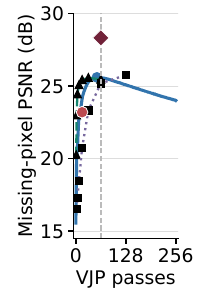}
\caption{Reconstruction}\label{fig:adam:quality}
\end{subfigure}%
\hspace{.006666\textwidth}%
\begin{subfigure}[t]{0.245\textwidth}
\centering\includegraphics[width=\linewidth]{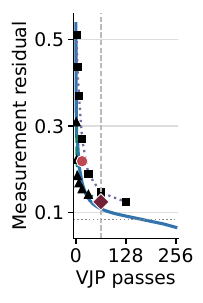}
\caption{Optimized loss}\label{fig:adam:residual}
\end{subfigure}%
\hspace{.006666\textwidth}%
\begin{subfigure}[t]{0.245\textwidth}
\centering\includegraphics[width=\linewidth]{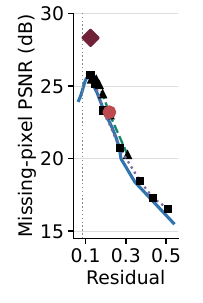}
\caption{Loss vs. quality}\label{fig:adam:tradeoff}
\end{subfigure}%
\hspace{.006666\textwidth}%
\begin{subfigure}[t]{0.245\textwidth}
\centering\includegraphics[width=\linewidth]{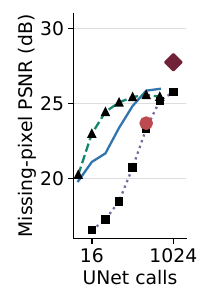}
\caption{Network cost}\label{fig:adam:compute}
\end{subfigure}%
}
\par\vspace{2pt}\includegraphics[width=\linewidth]{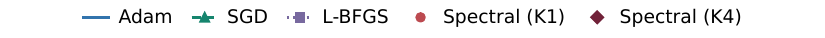}
\caption{\textbf{Residual minimization can degrade reconstruction.} (a--c) Adam and spectral use 50 paired reference images; SGD/L-BFGS use the separate 50-image optimizer set also in (d). Adam loses 1.59 dB after its peak while residual falls. Dashed guides mark 64 VJPs; dotted guides mark clean-image residual. L-BFGS uses recorded backward counts. (d) Log-scale UNet evaluations charge VJP and JVP traversals. Paired budget comparisons: Appendix~\ref{app:adam-budget}.}
\label{fig:adam-collapse}\label{fig:optimizer-main}
\end{figure}

\paragraph{Residual versus reconstruction.}
On the separate paired trajectory in Fig.~\ref{fig:adam-collapse}, Adam peaks at 25.60 dB near step 54, then loses 1.59 dB by step 256 despite decreasing measurement loss. K4 reaches 28.32 dB: it exceeds Adam@64 by 2.74 dB at equal VJP count and Adam@128 by 3.25 dB at equal total derivative traversals (95\% interval $[2.32,4.35]$; 86\% paired wins). K1 trails Adam under both conventions (Appendix~\ref{app:adam-budget}); the benefit here comes from repeated geometry correction.
\paragraph{Guidance and structured edits.}\label{sec:clamp-stress}
In Fig.~\ref{fig:vision-reconciled}(a--d), spectral exceeds selected Universal Guidance by 0.133 dB on super-resolution and 0.140 dB on deblurring, with no significant paired difference on inpainting or composite restoration; guidance obtains early gains at lower cost on three operators. At a $4\times$ structured-edit request, K1/K2 reach cosine 0.929/0.996 versus 0.845 for CLAMP. The stress test probes realization of an analytic property beyond reconstruction. These are descriptive objective adaptations; marginal intervals alone do not establish paired significance. Across the evaluated settings, spectral has no below-prior outcomes, while selected unstable guidance settings do. Implementation and sensitivity details are in Appendices~\ref{app:vision-populations} and~\ref{app:ugprotocol}.

\subsection{Audio: precise change and preservation}\label{sec:audio}
Across four practical mel-domain tasks, K2 attains 100\% joint success versus 8--82\% for FO. At 64 backward passes on 30 selective-EQ clips, K4 achieves 0.00038 dB all-coordinate RMSE versus Adam's 0.02997, winning all pairs. SGD reaches 0.00047 dB with no significant difference from K4 ($p=.63$); all three satisfy the success threshold. The precision comparison remains informative after the binary success threshold saturates: K4 and Adam both pass, but differ in realized coordinate errors. Transition edits further show that meeting the measured request can increase an unoptimized abruptness diagnostic. These are acoustic-control results, without a listening-study claim (Appendix~\ref{app:audio-detail}).

Waveform localized EQ reaches 2.691 dB error at K16 versus 3.895 for L-BFGS at the reported settings. Bandwidth extension is budget sensitive: Adam's error is 26.604 dB at 16 steps versus K4's 12.730, but early-stopped Adam reaches 12.525 with no significant paired difference. Table~\ref{tab:cross-domain} therefore establishes neither uniform superiority nor equal-cost advantages across audio tasks.

\subsection{Robotics: transferring corrections to execution}\label{sec:robot}
\begin{figure}[t]
\centering
\makebox[\textwidth][c]{%
\begin{subfigure}[t]{.18\textwidth}
\centering\parbox[c][129.6bp][c]{\linewidth}{\includegraphics[width=\linewidth]{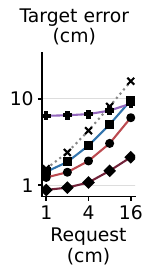}}
\caption{Reaching}\label{fig:rob:reach}
\end{subfigure}%
\hspace{.0025\textwidth}%
\begin{subfigure}[t]{.20\textwidth}
\centering\parbox[c][129.6bp][c]{\linewidth}{\includegraphics[width=\linewidth]{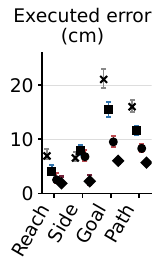}}
\caption{Specifications}\label{fig:rob:specs}
\end{subfigure}%
\hspace{.0025\textwidth}%
\begin{subfigure}[t]{.18\textwidth}
\centering\parbox[c][129.6bp][c]{\linewidth}{\includegraphics[width=\linewidth]{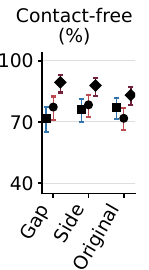}}
\caption{Obstacles}\label{fig:rob:obstacle}
\end{subfigure}%
\hspace{.0025\textwidth}%
\begin{subfigure}[t]{.19\textwidth}
\centering\parbox[c][129.6bp][c]{\linewidth}{\includegraphics[width=\linewidth]{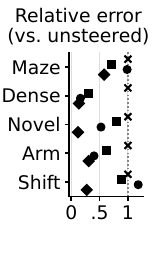}}
\caption{Planning}\label{fig:rob:planning}
\end{subfigure}%
\hspace{.0025\textwidth}%
\begin{subfigure}[t]{.24\textwidth}
\centering\parbox[c][129.6bp][c]{\linewidth}{\includegraphics[width=\linewidth]{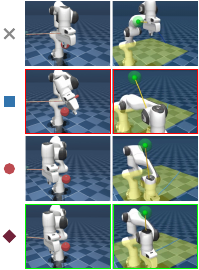}}
\caption{Manipulation}\label{fig:rob:example}
\end{subfigure}%
}
\par\vspace{2pt}\includegraphics[width=\linewidth]{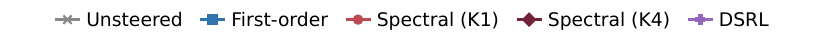}
\caption{\textbf{Robotics and planning outcomes.} (a,b) Executed error medians and 95\% intervals for Panda reaching, orientation/height preservation, and Cheetah goal/path tracking. (c) Contact-free execution rates. Unsteered fails the reach, so its contact-free score is omitted. (d) Endpoints normalized by unsteered (lower is better): Maze2D distance, dense/novel-obstacle collision, arm error, shifted-goal failure. Counts: Appendix~\ref{app:vision-populations}. (e) Paired obstacle-clearance (left) and constant-height (right) examples; symbols identify methods via the shared legend.}
\label{fig:rob}
\end{figure}

On 276 Panda instances, K1 reduces median executed reaching error from FO's 2.39 cm to 1.06 cm; learned DSRL reaches 6.01 cm after 8,000 environment episodes per seed. DSRL learns from black-box rewards; spectral uses differentiability at inference time, so their access and deployment conditions differ. Figure~\ref{fig:rob}(a--c,e) covers reaching, preservation, clearance, and paired manipulation examples.

Table~\ref{tab:cross-domain} shows benefits on Panda pose and Gymnasium HalfCheetah trajectory tracking~\citep{robomimic2021}. On shared reaching instances, K8 ties L-BFGS at 0.63 cm; K16's 0.59 cm is not a significant improvement. Greater clearance need not increase joint reach-and-clear success, and depth can be non-monotone. The unsteered contact-free score is omitted from the plot because it fails the reach. The qualitative rows pair obstacle clearance with a constant-height constraint. Generated fidelity predicts executed improvement more strongly once requests exceed servo resolution, reaching rank correlation 0.77; smaller corrections can be attenuated during tracking. Execution and refinement analyses are in Appendices~\ref{app:boundaries} and~\ref{app:vision-populations}.
\subsection{Control and planning}\label{sec:control}
\paragraph{Character motion generation.}\label{sec:character}
A frozen MDM produces three-second clips with 25 deterministic DDIM transitions remaining at intervention. On 50 paired seeds per task, avoidance ($Z=13$) and multi-objective reaching ($Z=77$) combine pelvis or wrist changes with preservation (Fig.~\ref{fig:method}; Appendix~\ref{app:motion}). K1 improves cosine over movement-matched FO from 0.458 to 0.976 and from 0.701 to 0.931, winning every pair. Relinearization reduces remaining positional error. At K4-matched nominal backward-pass counts, K4 achieves 96\% joint success on both tasks versus L-BFGS's 98\% and 96\%. On reaching, K4 limits bone drift to 1.05 cm versus 5.66 for SGD and 7.20 for Adam. Joint success checks positions and jerk; collision, velocity, and bone drift are separate diagnostics. For example, Adam satisfies the avoidance success gate in 88\% of cases but is collision-free in 78\%. This distinguishes coordination of measured motion properties from additional physical constraints. The comparison matches nominal backward passes, not standalone runtime. These results support structured motion control while leaving generic optimizers competitive.

\paragraph{Trajectory control and planning.}\label{sec:trajectory-control}
Across 45 executed waypoint pairs, K4 reduces Adam's mean error from 7.48 to 3.05 cm (59\%; 41/45 wins; paired difference interval $[2.96,5.86]$ cm), also improving on fixed GD and L-BFGS-B settings (Appendix~\ref{app:one-waypoint}). Figure~\ref{fig:rob}(d) shows shifted-goal success rising from FO's 26.5\% to K4's 77.4\%, although K1 reaches only 2.6\%; unseen-obstacle collision falls from 79.5\% to 11.8\%. Specialized guidance can achieve lower collision rates. The relative ranking depends on goal accuracy, clearance, and budget (Appendix~\ref{app:executed-guidance}).

\section{Discussion and Limitations}
The evidence supports correction of structured requests whose local model remains informative. Inverse corrections and nonlinear optimizers both refine residuals; their relative outcomes depend on task and budget, and relinearization need not improve monotonically. Coordinate scaling determines the geometry; omitted properties receive no protection. Generated fidelity can be attenuated or overwritten during execution and replanning. The predictor's pooled validation uses retrospective regime selection and concerns cosine alignment rather than task success. Derivative counts do not imply runtime gains. Robotics is simulated and audio has no listening study; the appendix analyzes these boundaries alongside optimizer counterexamples.

\paragraph{Toward precise downstream control.}
A natural direction is to use the local inverse as a precision layer within closed-loop diffusion policies, which already support receding-horizon execution~\citep{chi2023diffusionpolicy}. New observations would trigger a new geometry estimate and requests for tracking, clearance, or terminal-state changes. Differentiable dynamics and conservation residuals could encode physical-law targets. Such a layer would require execution-aware validation and independent safety enforcement: local request fidelity alone does not guarantee constraint satisfaction. A complementary direction replaces hand-designed properties with learned semantic features, requesting targeted edits while preserving selected attributes. Its value depends on feature calibration, local controllability, and whether the representation measures the intended semantics. Appendix~\ref{app:control-perspective} formulates these directions and their evaluation requirements.

\section{Conclusion}\label{sec:conclusion}
Spectral steering connects a structured request's modal composition to its distortion through a frozen generator, the predicted benefit of local inverse correction, and the realized outcome. Scalar nulls and matched-movement controls test this connection; cross-domain experiments establish its practical relevance, including complex character motion generation. Classical damped least squares supplies the correction, while fixed-time relinearization extends its use to finite edits. The evidence supports request-dependent geometry as a useful control interface, with benefits bounded by local-model accuracy, objective coverage, and downstream execution. Future work should further characterize how precise generator-space steering transfers to downstream tasks, including closed-loop control under physical and safety constraints and targeted semantic editing with learned representations, while deriving theoretical locality bounds on the maximum reliable steering magnitude.
\label{main:last}
\clearpage

\subsection*{AI use statement}
Generative AI tools were used for code assistance, result interpretation, and manuscript drafting/editing. The authors remain responsible for citation accuracy, originality, policy compliance, and the final manuscript.

\subsection*{Ethics statement}
This work studies inference-time steering of pretrained generators and inherits their misuse and deployment risks. Robotics experiments are simulator based and do not establish safety guarantees. Improved request fidelity is not guaranteed task success.

\subsection*{Reproducibility statement}
The supplement records objectives, checkpoints, matching protocols, optimizer budgets, paired statistics, figure source data, and available qualitative selection rules. The accompanying additional material supports the paper's claims with code and media files. Media files include image, video, and audio samples of the steered models. The appendix specifies implementation and evaluation protocols.

\bibliography{references}
\bibliographystyle{plainnat}

\clearpage
\appendix

\input{appendix_layout}
\section*{Appendix contents}
The appendix opens with supporting method and comparison details, followed by
additional quantitative analyses, comparison protocols, and solver details. The qualitative gallery then follows the main-table order:
vision, audio, robotics, and control and planning, with character motion generation
first in the last category. Task objectives and evaluation endpoints are collected
at the end. The entries below link to the corresponding sections.
\printappendixcontents
\clearpage
\setupappendixlayout
\section{Supporting Analysis and Comparisons}\label{app:main-details}
We give the scalar control and support diagnostic, then summarize the control interpretation and complementary domain findings. Detailed protocols and qualitative examples follow.

\paragraph{Local model and exact scalar null.}\label{sec:scalar-null}
Section~\ref{sec:method} gives the local model and primal/dual solves; the dual system has dimension $Z$ rather than $D$. For $Z=1$, write $\Jh=j^\top$ with $j\neq0$ and $d\neq0$. The inverse is $jd/(\|j\|_2^2+\lambda)$. Matching predicted objective-space movement sets FO's step size to $(\|j\|_2^2+\lambda)^{-1}$, making the latent updates, and hence the outputs under the same sampler randomness, identical. The 935-pair scalar experiment tests this exact null; Appendix~\ref{app:direction} tests the multi-coordinate case.

\paragraph{Damping and numerical support.}
The coefficient $\gamma=0.01$ is fixed; $\lambda_k=\gamma\sigma_1(J_k)^2$ is refreshed at each correction. For nonzero $d$, numerical reachability is
\begin{equation}
A_{\rm reach}(d)=\frac{\|U_rU_r^\top d\|_2^2}{\|d\|_2^2}.
\label{eq:reach}
\end{equation}
The SD1.5 diagnostic retains $\sigma_i>10^{-4}\sigma_1$ without truncating the damped correction. Measured real-task requests were fully reachable, so natural rank deficiency is not established here. If $J=0$, no first-order response is available. Appendix~\ref{app:solver-implementation} specifies solver variants, thresholds, and stopping rules.

\paragraph{Local inversion as a control interface.}\label{app:control-perspective}
The Jacobian maps perturbations of the full intermediate state to selected properties; its damped inverse maps a request back to an intervention. ``Direct'' describes this inverse relation: matrix-free solves and finite-edit relinearization still iterate. A future precision layer could re-estimate this geometry from each new observation within a receding-horizon diffusion policy~\citep{chi2023diffusionpolicy}, using tracking, clearance, energy, or dynamics residuals as properties. For example,
\[
h(y)=\big[p_{\rm endpoint}(y),\ r_{\rm dynamics}(y),\ E(y),\ c_{\rm clearance}(y)\big]^\top.
\]
A zero dynamics target requires $d_{\rm dynamics}=-r_{\rm dynamics}(y)$; zero displacement instead preserves the current residual. Execution-aware validation and independent safety enforcement remain necessary. Learned semantic properties offer a complementary direction, with edit and preservation quality evaluated independently of the optimized features. Neither extension is evaluated here.

\paragraph{Table conventions.}
Bold marks numerical winners, including rounded ties. The two ns-marked methods in a row have no reported statistically significant difference; this does not establish equivalence. The structured-edit unsteered cosine of 0.000 is a display convention for an undefined zero-edit cosine. Qualitative selection rules appear in Appendix~\ref{app:qualitative-gallery}.

\paragraph{Vision.}
The selected-guidance comparisons give spectral advantages of 0.133 dB on super-resolution ($p=4\times10^{-8}$) and 0.140 dB on deblurring ($p=.002$). Appendix~\ref{app:ugprotocol} reports the setting-specific instability counts, and Appendix~\ref{app:adam-budget} gives the paired optimizer comparisons.

\paragraph{Audio.}
K1 raises localized mel-EQ success from 8\% to 94\% and paired-clip consistency from 25\% to 100\%. On 30 selective-EQ clips at 64 backward passes, K4/Adam/SGD latent displacements are 7.85/19.03/7.02. Appendix~\ref{app:audio-detail} reports coordinate errors and the unoptimized abruptness diagnostic; these are acoustic-control measurements, not listening-study outcomes.

\paragraph{Robotics.}
DSRL's 6.01 cm reaching error improves its separate 23.6 cm training reference after 8,000 episodes per seed over five training seeds; spectral's unsteered reference is 3.89 cm. DSRL amortizes learned adaptation, while spectral differentiates at inference time. Figure~\ref{fig:rob:transfer} shows how request scale affects transfer to execution. A via-point example improves generated cosine by 0.0674 without detectable executed waypoint improvement ($p=.77$).

\paragraph{Control and planning.}
After K1, median character errors are 2.92 cm for avoidance and 2.21 cm for reaching, above their 1.66 cm and 1.46 cm tolerances. Directional fidelity therefore need not complete the request (Appendix~\ref{app:motion}). For trajectory control, Appendix~\ref{app:one-waypoint} gives paired executed errors; Appendix~\ref{app:executed-guidance} compares clearance and goal accuracy. Dropping Panda clearance coordinates reduces target error from 1.30 to 0.70 cm while lowering contact-free execution from 83.2\% to 61.9\%, illustrating the importance of objective coverage (Appendix~\ref{app:boundaries}).

\paragraph{Connection to prior control and guidance.}
End-to-end optimization, including D-Flow~\citep{benhamu2024dflow}, establishes full-process differentiation and output-gradient geometry; damped least squares and relinearization are classical (Section~\ref{sec:related_work}). Here the predictor, scalar null, and movement controls test how request--geometry interactions explain realized directional gains. Finite-edit and execution results test their practical consequences. The predictor requires derivative information and is not a certified step-acceptance rule.

\section{The Correction Is a Direction}\label{app:direction}

\begin{figure}[!htbp]
\centering
\begin{subfigure}[b]{0.485\textwidth}\includegraphics[width=\linewidth]{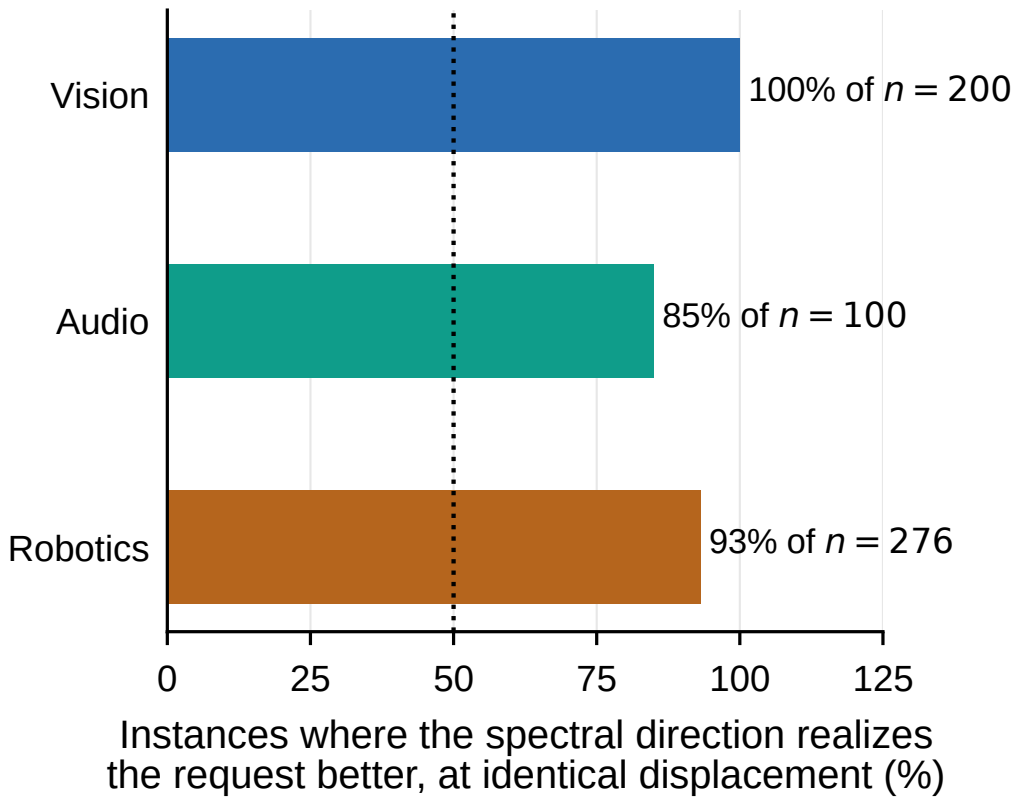}
\caption{Exactly matched latent displacement}\label{fig:dir:a}\end{subfigure}\hfill
\begin{subfigure}[b]{0.485\textwidth}\includegraphics[width=\linewidth]{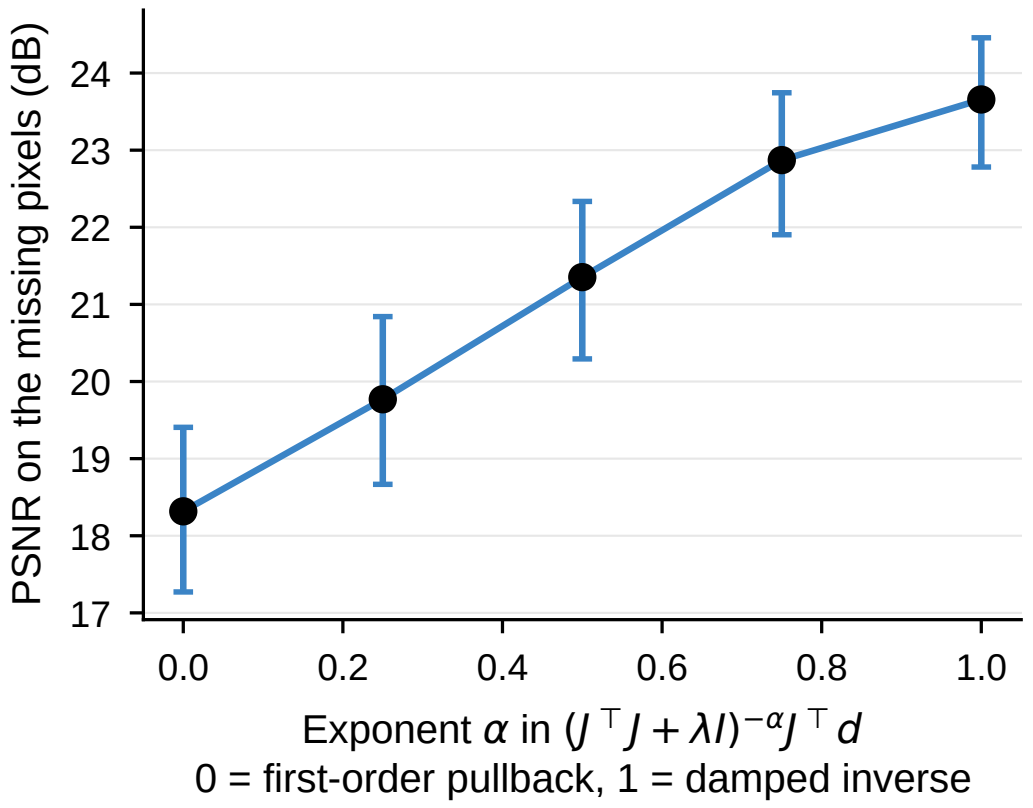}
\caption{Interpolating the direction alone}\label{fig:dir:b}\end{subfigure}
\caption{(\subref{fig:dir:a}) With latent-displacement norms agreeing within $4\times10^{-14}$, the spectral direction
realizes the request better on almost every instance in three modalities. (\subref{fig:dir:b}) The
fractional family $\dx(\alpha)=(\Jh^\top\Jh+\lambda I)^{-\alpha}\Jh^\top d$ at exactly matched
\emph{predicted objective-space} magnitude $\|\Jh\dx\|_2$ interpolates from the first-order pullback ($\alpha=0$) to the damped inverse
($\alpha=1$) and moves reconstruction monotonically, improving on 50 of 50 images at every step. Neither
panel is a compute comparison; both vary only the direction.}
\label{fig:dir}
\end{figure}

The scalar null in Appendix~\ref{sec:scalar-null} predicts identical updates
for a single property. The multi-coordinate controls here test whether spectral
steering benefits from its update direction after accounting for movement magnitude. Figure~\ref{fig:dir} first fixes latent
displacement, then varies the direction at fixed predicted downstream movement.
The two controls separate geometry correction from simply taking a larger step.

\section{Failure Modes and Scope}\label{app:boundaries}

\begin{figure}[!htbp]
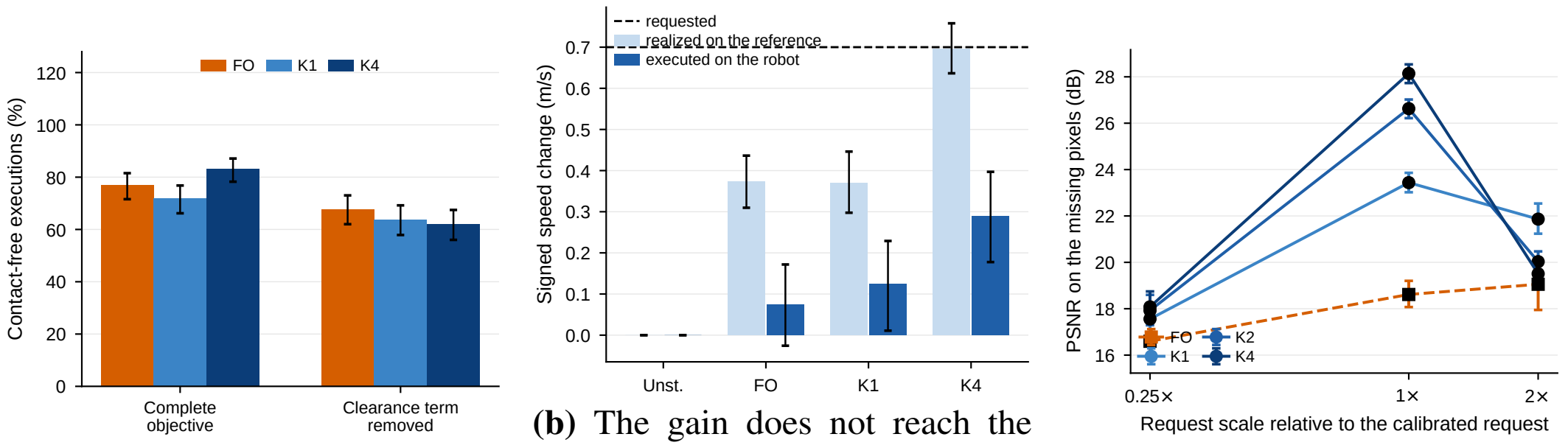

\centering
\begin{subfigure}[b]{0.32\textwidth}\includegraphics[width=\linewidth]{\panel{p_bound_incomplete}}
\caption{An incomplete objective}\label{fig:bnd2:c}\end{subfigure}\hfill
\begin{subfigure}[b]{0.32\textwidth}\includegraphics[width=\linewidth]{\panel{p_bound_transmission}}
\caption{The gain does not reach the robot}\label{fig:bnd2:a}\end{subfigure}\hfill
\begin{subfigure}[b]{0.32\textwidth}\includegraphics[width=\linewidth]{\panel{p_bound_overrequest}}
\caption{The request itself is wrong}\label{fig:bnd2:b}\end{subfigure}
\caption{(\subref{fig:bnd2:c}) Dropping the clearance coordinate improves the executed target error and
lowers contact-free execution, and the strongest rung loses the most: a more faithful inverse of an
incomplete request is a worse outcome. (\subref{fig:bnd2:a}) A signed speed request is realized on the
generated reference at cosine 0.998 and the robot executes 41\% of what was asked; generated and
executed quantities are separate bars because they are separate measurements. (\subref{fig:bnd2:b}) At a
doubled request the deeper rungs pursue the overshoot faithfully and finish below one solve.}
\label{fig:bnd2}
\end{figure}

\begin{figure}[!htbp]
\centering
\begin{minipage}[c]{.245\textwidth}
\includegraphics[width=\linewidth]{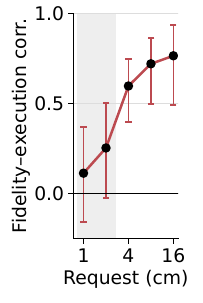}
\end{minipage}\hfill
\begin{minipage}[c]{.71\textwidth}
\caption{\textbf{When generated fidelity transfers to robot execution.}
For Panda reaching, Spearman correlation between the K1-minus-FO gain in generated
request cosine and the FO-minus-K1 reduction in executed target error, within each request size.
Error bars show 95\% bootstrap intervals ($n=50$--59 per size).
The correlation rises from 0.11 at 1 cm to 0.77 at 16 cm; shading marks requests
below the servo's resolution. This scale dependence complements the transmission
failure in Fig.~\ref{fig:bnd2}\subref{fig:bnd2:a}.}
\label{fig:rob:transfer}
\end{minipage}
\end{figure}

Local request fidelity is only one requirement for useful steering.
Figure~\ref{fig:bnd2} separates incomplete specifications, loss of a correction
during execution, and an excessive requested change. The transfer analysis and
optimizer counterexample below identify further limits on the main-paper gains.

\paragraph{A complete optimizer counterexample.}
On the separate frozen dense-obstacle B2 task ($n=60$), tuned optimizers outperform
the spectral ladder on the preregistered request residual. At about 128 backward
passes, SGD and L-BFGS achieve mean residuals 0.0446 and 0.0422 versus K4's
0.6364; L-BFGS also has higher mean minimum clearance (0.09988 versus 0.09294).
Thus the executed waypoint advantage does not establish a general planning
advantage. An exploratory path-length difference in this campaign is not used
to reverse its primary-endpoint conclusion.

Across these cases, useful steering requires objective coverage, a valid local
model, and reliable downstream execution.

\section{Additional Domain Analyses}\label{app:additional-plots}

This section supplements the main results with acoustic precision and locality
analyses, followed by the pooled directional-prediction audit. It distinguishes
an accurately realized request from independent quality criteria and documents
how the local-regime population is selected.

\subsection{Practical audio protocol and remaining boundaries}\label{app:audio-detail}
The practical audio campaign uses one frozen mel-domain prior with differentiable
time--frequency and loudness coordinates. At the headline request, localized EQ,
two-clip transitions, reference tonal matching, and paired-clip consistency contain
50, 50, 50, and 40 paired instances. All coordinates must meet the campaign's
predefined tolerances. The localized-EQ optimizer comparison uses the first 30
instances of request A1c, with three change and thirteen preservation coordinates.
The selective-EQ row in Table~\ref{tab:cross-domain} reports the mean of the per-instance RMSE over all sixteen coordinates,
not a mean of the two block RMSEs. Adam uses learning rate 0.01 and SGD 0.1, selected using 32-step runs on five
disjoint pilot seeds (900--904), then frozen for all evaluation budgets. Budgets of 16, 32, and 64 backward
passes correspond to K1, K2, and K4 in this explicit-Jacobian audio implementation;
these counts are not wall-clock matches.

\begin{figure}[!htbp]
\centering
\includegraphics[width=\linewidth]{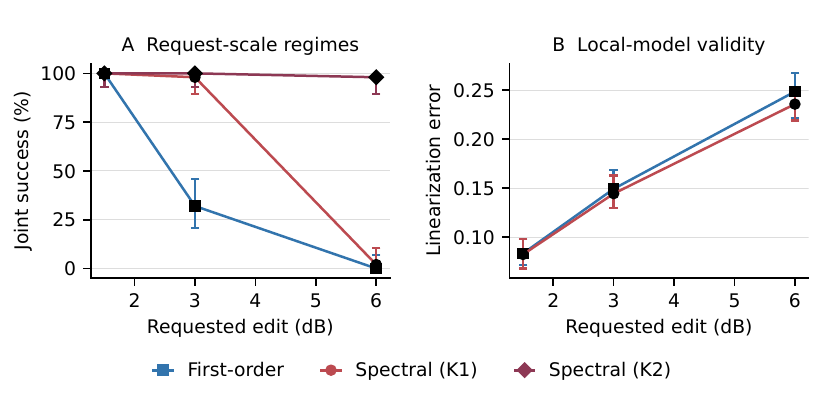}
\caption{\textbf{Audio request scale separates direction from locality.}
At the calibrated edit, one solve improves success; at twice the request,
relinearization is needed. FO and spectral incur similar linearization error
at matched scales (16-cell diagnostic), so their intermediate-scale difference
is directional. These scale sweeps complement the practical tasks in the main text.}
\label{fig:aud}
\end{figure}

At 64 passes, change / preservation RMSE is 0.000308 / 0.000392 dB for spectral,
0.034511 / 0.028056 for Adam, and 0.000413 / 0.000477 for SGD.
The mean all-coordinate error reduction relative to Adam@64 is 0.02958 dB
(95\% paired bootstrap interval [0.02760, 0.03144], 30/30 wins;
two-sided Wilcoxon $p=1.86\times10^{-9}$). Against SGD@64 the interval includes
zero ($p=.63$). SGD wins at smaller budgets and nearly matches spectral at 64 passes; no significant
spectral advantage over SGD is claimed. On transitions, every steered variant
increases unoptimized energy abruptness from 9.47 to 11.3--11.6 dB.
Separate waveform-prior tests show that one solve does not improve bandwidth
extension and that preservation requires coordinates that actually isolate the
held region: waveform localized EQ worsens pinned cells by 0.0834 dB at K4.
These outcomes do not imply a listening-study or naturalness benefit. Figure~\ref{fig:aud} examines how request scale changes success and local-model accuracy.

\subsection{Predictor scope}\label{app:predictor-audit}

\begin{figure}[!htbp]
\centering
\includegraphics[width=.70\linewidth]{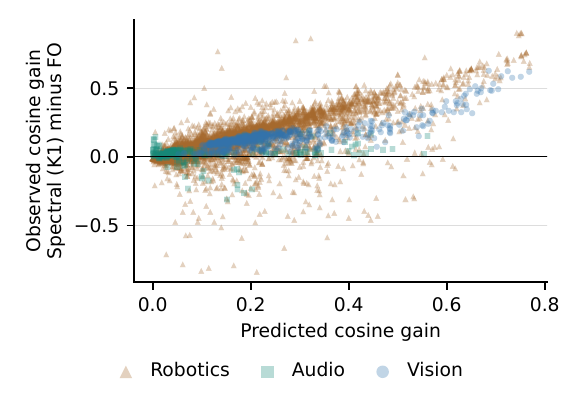}
\caption{\textbf{Cross-domain prediction of directional benefit.}
Each point is one of 8,923 instances from 27 vision, audio, and robotics
populations in the measured local regime. The horizontal axis is the pre-edit
predicted cosine gain; the vertical axis is the observed cosine gain of K1
over magnitude-matched FO ($r=0.75$, $\rho=0.79$). Positive vertical values
favor K1. Population-level local-regime selection uses nonlinear responses
and is retrospective. The motion illustration in Fig.~\ref{fig:mech}\subref{fig:mech:predictor}
is a separate diagnostic population and is not included here.}
\label{fig:pred}
\end{figure}

\begin{figure}[!htbp]
\centering
\includegraphics[width=\linewidth]{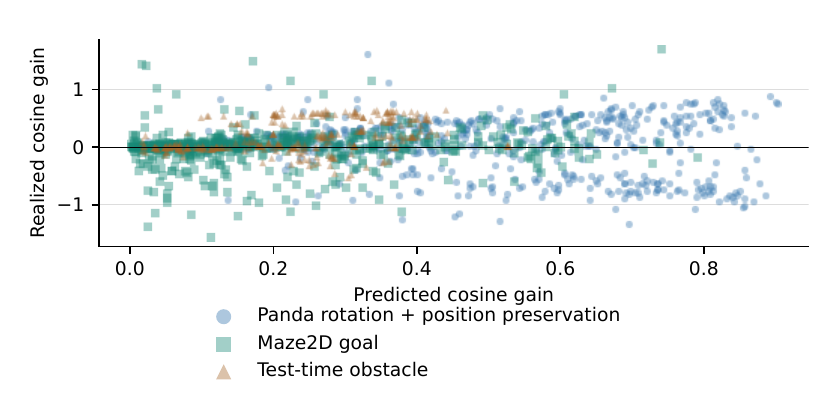}
\caption{\textbf{The local-model boundary.} The 1,439 instances from the three
populations excluded by the retrospective linearization-error criterion.
Their broad realized responses show why the headline local-regime correlation
should not be extrapolated to large nonlinear edits.}
\label{fig:predictor-outside}
\end{figure}

We compare the predicted and realized gain in request cosine, then examine where
the local model ceases to be informative. Figure~\ref{fig:pred} summarizes the
retained populations; Fig.~\ref{fig:predictor-outside} shows the excluded
nonlinear regimes. Selection is retrospective, as detailed below.

The complete diagnostic collection contains 30 populations and 10,362 instances.
Three populations have median one-shot linearization error above one:
Panda rotation with position preservation, Maze2D goal reaching, and the
test-time obstacle. Removing those leaves the 27 populations and 8,923 instances
in Fig.~\ref{fig:pred}, with positive Pearson correlation in 26 populations.
After centering within each population, the pooled correlations remain
Pearson 0.74 and Spearman 0.77. Regime assignment uses measured nonlinear
responses and is retrospective, so these results do not establish an automatic
pre-edit acceptance rule.

On a separate composite-restoration evaluation ($n=150$), the predictor
correlates positively with realized cosine gain ($r=0.805$, $\rho=0.766$),
but negatively with change-coordinate error reduction ($r=-0.438$,
$\rho=-0.424$). Directional fidelity and error reduction are distinct endpoints;
neither establishes independent reconstruction quality.

\section{Control and Planning: Character Motion Generation}\label{app:motion}

\paragraph{Population and generator.}
We report results on 50 seeds per task, using the same seeds across
methods. Comparator hyperparameters were selected on six separate
calibration instances, and the jerk threshold was calibrated on
24 additional generator samples. The frozen text-conditioned MDM~\citep{tevet2023human}
generator produces 60 frames at 20 fps using a native 50-step
sampling schedule and classifier-free guidance 2.5. Steering acts
on an intermediate HumanML3D feature state of dimension
$263\times1\times60=15{,}780$, with 25 deterministic DDIM
transitions remaining~\citep{song2020denoising}. All local corrections act at this same
diffusion time.

The main table reports joint success under the positional-target,
preserved-position, and jerk criteria defined below, whereas
Table~\ref{tab:motion-ladder} reports target-only success.
These criteria must not be interchanged. The main-table entries
for SGD, Adam, and L-BFGS use the K4-matched nominal backward-pass
count; this does not match forward evaluations or standalone runtime.

\paragraph{Solver and first-order control.}
We differentiate the scaled objectives through the complete remaining
sampler and differentiable joint reconstruction. Each correction uses
the explicit Jacobian with reduced-SVD filtering and
$\lambda_k=0.01\sigma_1(J_k)^2$, recomputed at each relinearization
toward the fixed target $z(x_t^{(0)})+d$.
FO is analytically scaled to match K1's predicted objective-space
displacement norm.

\paragraph{Requests and preservation coordinates.}
Directions are expressed in a subject-centred frame measured from each baseline motion. The requested magnitude is one natural excursion of that motion: 16.6 cm of lateral pelvis travel for avoidance and 14.6 cm of wrist-relative-to-pelvis travel for reaching. Obstacle avoidance uses $Z=13$: a 20 cm sphere is placed on the body path, the mid-clip pelvis is moved far enough to include a 25\% clearance margin, and the final pelvis plus seven arm-bone lengths are preserved. Multi-objective reaching uses $Z=77$: the wrist follows a five-frame profile, both feet are pinned at two additional frames, a zero wrist-velocity target is imposed on arrival, and bone lengths are preserved at all waypoint frames.
Joint success requires target error below $0.1a$, preserved-position
error below $0.2a$, and mean third-order joint-position differences
below $0.0417139464$ m, computed without time-step normalization.
Here, $a=0.16636$ m for avoidance and $a=0.14647$ m for reaching.
Target error is the pelvis-position error at the avoidance waypoint
or the mean wrist-target distance over the five reaching waypoints.
Preservation error is the maximum displacement over preserved
joint--frame positions.
Position and bone-length coordinates are divided by $0.10$ m,
and velocity coordinates by $2.0$ m/s before constructing the
objective and its Jacobian.

\begin{table}[!htbp]
\centering\small
\caption{\textbf{Character-animation relinearization ladder} ($n=50$). Target error is averaged over the 50 seeds and reported in cm; success is the fraction within 10\% of the requested amplitude. $E_{\mathrm{lin}}$ measures deviation of the realized change from the tangent prediction.}
\label{tab:motion-ladder}
\begin{tabular}{@{}llrrrrrr@{}}
\toprule
Task & Endpoint & Unst. & FO & K1 & K2 & K4 & K16 \\
\midrule
\multirow{3}{*}{Obstacle} & error $\downarrow$ & 16.64 & 10.32 & 3.54 & 0.95 & 0.23 & \textbf{0.01} \\
 & success \% $\uparrow$ & -- & -- & 18 & 88 & 98 & \textbf{100} \\
 & $E_{\mathrm{lin}}\downarrow$ & -- & -- & 0.164 & 0.140 & 0.135 & \textbf{0.132} \\
\addlinespace
\multirow{3}{*}{Multi-objective} & error $\downarrow$ & 10.62 & 4.85 & 2.52 & 1.40 & 0.79 & \textbf{0.19} \\
 & success \% $\uparrow$ & -- & -- & 14 & 58 & 96 & \textbf{98} \\
 & $E_{\mathrm{lin}}\downarrow$ & -- & -- & 0.188 & 0.204 & 0.210 & \textbf{0.185} \\
\bottomrule
\end{tabular}
\end{table}

Table~\ref{tab:motion-ladder} shows how relinearization reduces
positional target error across both tasks.

A separate computational benchmark on the $Z=77$ task reports
8.61 s for 77 optimizer steps, compared with 3.80 s for sequential
Jacobian-row construction and 0.95 s for vectorized construction.
The Jacobian rows share one forward graph, whereas optimizer steps
evaluate successive states. These component timings characterize
Jacobian-construction efficiency; they do not measure the complete
standalone runtime of K4.

\begin{table}[!htbp]
\centering\small
\caption{\textbf{One spectral solve versus magnitude-matched FO} ($n=50$, paired). Both methods start from the same $x_t$ and request and are matched in predicted objective-space displacement norm. Leakage is unintended movement relative to requested movement.}
\label{tab:motion-paired}
\setlength{\tabcolsep}{3pt}
\begin{tabular}{@{}llrrrrrr@{}}
\toprule
Task & Quantity & FO & K1 & $\Delta$ & 95\% CI & W/L & $p$ \\
\midrule
\multirow{3}{*}{Obstacle} & fidelity cosine $\uparrow$ & 0.458 & \textbf{0.976} & $+0.519$ & $[+0.492,+0.546]$ & 50/0 & $2\!\times\!10^{-15}$ \\
 & leakage $\downarrow$ & 1.215 & \textbf{0.219} & $-0.997$ & $[-1.073,-0.918]$ & 50/0 & $2\!\times\!10^{-15}$ \\
 & bone drift (m) $\downarrow$ & \textbf{0.011} & \textbf{0.011} & $+0.000$ & $[-0.000,+0.000]$ & 27/23 & 0.67 (ns) \\
\addlinespace
\multirow{3}{*}{Multi-objective} & fidelity cosine $\uparrow$ & 0.701 & \textbf{0.931} & $+0.229$ & $[+0.216,+0.242]$ & 50/0 & $2\!\times\!10^{-15}$ \\
 & leakage $\downarrow$ & 1.477 & \textbf{0.507} & $-0.970$ & $[-1.040,-0.894]$ & 50/0 & $2\!\times\!10^{-15}$ \\
 & bone drift (m) $\downarrow$ & 0.011 & \textbf{0.010} & $-0.000$ & $[-0.000,-0.000]$ & 27/23 & 0.67 (ns) \\
\bottomrule
\end{tabular}
\end{table}

\begin{figure}[!htbp]
\centering
\includegraphics[width=\linewidth,draft=false]{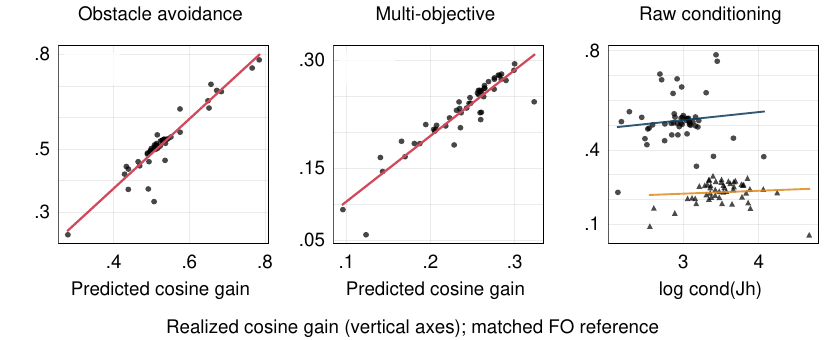}
\caption{\textbf{The request-specific predictor transfers to character animation.} $G_{\mathrm{cos}}^{\mathrm{pred}}$, computed before editing, predicts the realized cosine gain of K1 over magnitude-matched FO on obstacle avoidance and multi-objective reaching (Pearson $r=0.94$ and $0.92$; Spearman $\rho=0.92$ for both; $n=50$ seeds per task). Raw conditioning does not predict that gain within either task (right; $\rho=+0.04$ and $+0.18$). Correlations are computed within each task to avoid between-task offsets. The endpoint is cosine gain.}
\label{fig:motion-predictor}
\end{figure}

Figure~\ref{fig:motion-predictor} reports character-motion
correlations separately from the 27-population aggregate.

\begin{table}[!htbp]
\centering\small
\caption{\textbf{Deep refinement and adapted guidance} ($n=50$). A normalized residual above one means the result is farther from $h+d$ than the unsteered baseline.}
\label{tab:motion-fidelity}
\begin{tabular}{@{}lrrrr@{}}
\toprule
& \multicolumn{2}{c}{Obstacle} & \multicolumn{2}{c}{Multi-objective} \\
\cmidrule(lr){2-3}\cmidrule(lr){4-5}
Method & cosine $\uparrow$ & residual $\downarrow$ & cosine $\uparrow$ & residual $\downarrow$ \\
\midrule
Spectral (K16) & \textbf{1.000} & \textbf{0.001} & \textbf{0.996} & \textbf{0.044} \\
DPS-adapted & 0.290 & 1.605 & 0.575 & 1.777 \\
UG-adapted & 0.435 & 1.466 & 0.325 & 1.619 \\
\bottomrule
\end{tabular}
\end{table}

Table~\ref{tab:motion-fidelity} contrasts deep refinement with adapted guidance. The normalized residuals of matched FO and K1 are respectively 0.972 and 0.260
on avoidance, and 0.738 and 0.379 on multi-objective reaching. Their paired cosine
and leakage comparisons appear once, in Table~\ref{tab:motion-paired}.

\paragraph{Intervention time and guidance adaptation.}
Moving the obstacle intervention from 25 remaining denoising steps to only two improves the normalized residual of DPS-adapted from 1.605 to 0.930 and Universal Guidance from 1.466 to 0.437, but conjunctive success remains 0\% for DPS and rises only to 8\% for Universal Guidance. K4 remains at 96--98\% over the same sweep, while K1 degrades from 18\% to 0\%, exposing its sensitivity to a single local model.
Universal Guidance operates in the generator's clean motion-feature
space. The reported configuration uses five backward-guidance steps
and two self-recurrences, for 300 backward passes.

\section{Paired Adam Budget Audit}\label{app:adam-budget}

The same 50 inpainting cases support two complementary compute comparisons:
one matches VJP counts and the other counts both VJPs and JVPs.
Table~\ref{tab:adam-paired} reports reconstruction differences with paired
uncertainty. Adam leads one spectral solve at these budgets, whereas K4 leads
under both conventions; neither convention establishes equal wall-clock cost.

\begin{table}[!htbp]
\centering\small
\caption{\textbf{Paired inpainting reconstruction versus Adam at two resource-accounting conventions} ($n=50$). The difference is Spectral minus Adam on missing-pixel PSNR; positive values favor Spectral. Each spectral solve costs 16 VJPs and 16 JVPs, while each Adam iteration costs one VJP. Intervals are paired BCa 95\% intervals. The K4 leads survive Holm correction; bold marks the winner in each row.}
\label{tab:adam-paired}
\setlength{\tabcolsep}{3pt}
\begin{tabular}{@{}llrrr@{}}
\toprule
Accounting & Paired methods & $\Delta$ PSNR dB & 95\% interval & Spectral win \%\\
\midrule
VJP count & Spectral (K1) vs \textbf{Adam@16} & $-0.96$ & $[-1.45,-0.51]$ & 26\\
 & \textbf{Spectral (K4)} vs Adam@64 & \textbf{$+2.74$} & $[+1.78,+3.89]$ & 78\\
Total traversals & Spectral (K1) vs \textbf{Adam@32} & $-2.07$ & $[-2.73,-1.44]$ & 18\\
 & \textbf{Spectral (K4)} vs Adam@128 & \textbf{$+3.25$} & $[+2.32,+4.35]$ & 86\\
\bottomrule
\end{tabular}
\end{table}

\section{Comparison Protocols}\label{app:vision-populations}

This section specifies sample sets, optimizer selection, and aggregation for
the main comparisons, in table order: vision, audio, robotics, and control and
planning, beginning with character motion generation.

\paragraph{Vision restoration.}
The inpainting table uses 50 paired images (760--809); the guidance and optimizer-trajectory figures use images 200--249, with a spectral ladder on each set. On the table's pairs, SGD/Adam/L-BFGS minus K4 PSNR is $-2.18/-1.77/-1.99$ dB, with respective 95\% intervals $[-3.31,-1.36]$, $[-2.84,-1.02]$, and $[-3.04,-1.15]$. For SR, deblurring, and composite restoration, SGD/Adam use 50 images and L-BFGS uses 20. L-BFGS selects the best evaluation PSNR over $1,2,4,8,16,32,64$ iterations, choosing 16/64/64 (32/128/128 backward calls).

\paragraph{Structured-edit optimizer row.}
SD1.5 uses the standard-scale box-average color request (Objective B, window 10), seeds 6000--6019, and realized-request cosine. GD/Adam use pilot-frozen ten-step runs at rates 100/0.01; spectral uses refreshed relative damping and peaks at K8. L-BFGS reaches 0.765 at ten backward passes (five strong-Wolfe iterations), with the same regularized loss as GD/Adam,
$\tfrac12\|h(x+\dx)-h^\star\|_2^2+\tfrac12 10^{-3}\|\dx\|_2^2$.
Its rate 1.0 is evaluation-selected; rates 0.1/0.3 are ineligible because each stalls on one case. Paired tests use in-process GD/Adam reruns; the table retains their frozen summaries. Figure~\ref{fig:vision-reconciled}(e) uses $4\times$ requests.

\paragraph{Structured-edit guidance.}
The $4\times$ stress test uses objective adaptations. CLAMP implements the
published equations with stand-in constants, without validation against an
official code example. The displayed marginal intervals do not establish
paired significance; spectral collateral change was not measured on the same pass.

\paragraph{Waveform audio and budget sensitivity.}
SGD/Adam use 25 shared excerpts (100--124), with rates selected on the tuning objective using two disjoint excerpts. Table~\ref{tab:cross-domain} reports 16-step results. Bandwidth extension evaluates high-band reconstruction LSD, not the optimized measurement loss. Selecting the best evaluation LSD over $1,2,4,8,16$ steps yields SGD@1 (rate 0.3) and Adam@2 (rate 0.01). Adam@2 reaches 12.525 dB versus K4's 12.730, with paired K4-minus-Adam interval $[-0.44,0.83]$ dB and $p=.37$. This stopping-budget sensitivity test does not compare the table's Adam@16 cell.

Localized EQ reports requested-region error (\texttt{on\_err\_db}). SGD/Adam use rates $3\times10^{-6}/3\times10^{-4}$; SGD's near-zero step reflects its tuning rule. The completed row includes L-BFGS at 3.895 dB and K16 at 2.691 dB. Appendix~\ref{app:audio-detail} describes the separate mel-audio protocol.

\paragraph{Robotics refinement depth.}
On shared reaching instances, K8 and L-BFGS both have median error 0.63 cm ($p=1.00$); K16 reaches 0.59 cm without a significant paired difference. HalfCheetah target error worsens at K8 before recovering at K16. On obstacle reaching, K4/K8/K16 joint success is 75.8/75.5/76.2\% ($p=1.00$ for K4 versus K16), while contact-free execution increases from 83.2\% at K4 to 89.4\% at K16. Panda depth tests are paired within one re-execution.

\paragraph{Character motion generation.}
The character comparison uses 50 seeds per task, with separate calibration
instances and nominal backward-pass accounting. Appendix~\ref{app:motion}
defines its requests, matching rule, success criteria, and solver.
Target-only ladder success and the main table's joint success are distinct
reported endpoints.

\paragraph{Executed one-waypoint control.}\label{app:one-waypoint}
The additional table row reanalyzes the recorded executed waypoint endpoint from
the optimization extension: three frozen missions, fifteen seeds each, with all
methods on exactly the same 45 pairs. A six-coordinate request specifies the
timed waypoint, terminal position, and terminal velocity at a frozen diffusion
intervention. The SGD column is full-batch GD (30 steps, rate 0.5); Adam uses
40 steps at rate 0.3. The L-BFGS column is SciPy L-BFGS-B with no variable bounds,
100 maximum iterations and 300 maximum objective/gradient evaluations.
All three minimize the squared nonlinear request error plus
$10^{-3}\|\delta x\|_2^2$; their settings are fixed by the source protocol,
without a reported learning-rate sweep. Spectral uses the same frozen target
with its damped local solves. The recorded count for K4 is 28 Jacobian-building
forward/backward calls; the L-BFGS-B objective/gradient count averages 85.3.
These counters combine different operations and do not establish a runtime ratio.

Against GD, Adam, and L-BFGS-B, the paired mean executed-error reductions are
34.65, 4.43, and 23.30 cm; the respective percentile bootstrap 95\% intervals
are [29.46, 40.09], [2.96, 5.86], and [13.38, 34.49] cm
(20,000 paired resamples). All three two-sided Wilcoxon tests remain below
$10^{-8}$ after Holm correction over these comparisons. The uncertainty is
conditional on the three tested missions and chosen settings.
Generated task success is 100\% for both K4 and L-BFGS-B, whereas executed
success is 84.4\% and 53.3\%; Adam executes successfully on 71.1\%.
This distinction motivates reporting execution rather than generated success.

\paragraph{Planning endpoints in Figure~\ref{fig:rob:planning}.}
Maze2D uses 553 available evaluations for unsteered, FO, and K1, and 501 for K4.
Dense-obstacle and novel-obstacle collision rates use the paired reruns with 315 and 263
executions, respectively. Planar-arm error uses 380 evaluations per method, and shifted-goal
success uses 1,440. The plot divides distance, error, or collision rate by the unsteered
value within each task; shifted-goal success is first converted to failure rate.
Thus one denotes the unsteered endpoint and smaller values are better, without averaging
incompatible physical units. The accompanying source data retain all raw endpoints and counts.

\section{General Guidance on Executed Obstacles}\label{app:executed-guidance}

This planning comparison evaluates general guidance on the same simulator
executions as spectral steering. Table~\ref{tab:executed-guidance} reports
collision and goal-request error separately, because the multi-coordinate
objective can trade off clearance against the requested goal change.

\begin{table}[!htbp]
\centering\small
\caption{Novel test-time obstacle, $n=263$ paired executions. Collision is physical; request error scores the goal move in generator space. Passes are campaign backward-pass units. Bold marks the best displayed number in each endpoint column; unsteered has the lowest request error here.}
\label{tab:executed-guidance}
\begin{tabular}{lrrr}
\toprule
Method & Passes & Collision \% $\downarrow$ & Request error $\downarrow$\\
\midrule
\textbf{Unsteered} & 0 & 100.0 & \textbf{0.350}\\
FO & 45 & 79.5 & 0.522\\
Spectral (K1) & 45 & 52.5 & 0.677\\
Spectral (K4) & 180 & 11.8 & 0.417\\
Universal Guidance, fixed $m=1$ & 16 & 9.5 & 0.756\\
\textbf{Universal Guidance, fixed $m=16$} & 256 & \textbf{1.1} & 0.442\\
\bottomrule
\end{tabular}
\end{table}
On the dense-clearance condition ($n=315$), K4 and the best guidance setting
record 1.6\% and 0.0\% collisions respectively; the gap is small at this
low base rate. The novel-obstacle condition supplies the discriminating
comparison. Unsteered has lower median goal error than K4 here (0.350 versus 0.417),
so the relative K4--guidance request advantage is not an absolute goal-accuracy gain.
The 45-coordinate objective contains two goal coordinates, with the remainder
for clearance and preservation. Request errors are heavy-tailed; comparisons
use medians and paired win rates.

\section{Optimizer Trajectories and Resource Accounting}\label{app:optimizers}

\begin{figure}[!htbp]
\centering
\includegraphics[width=.62\linewidth]{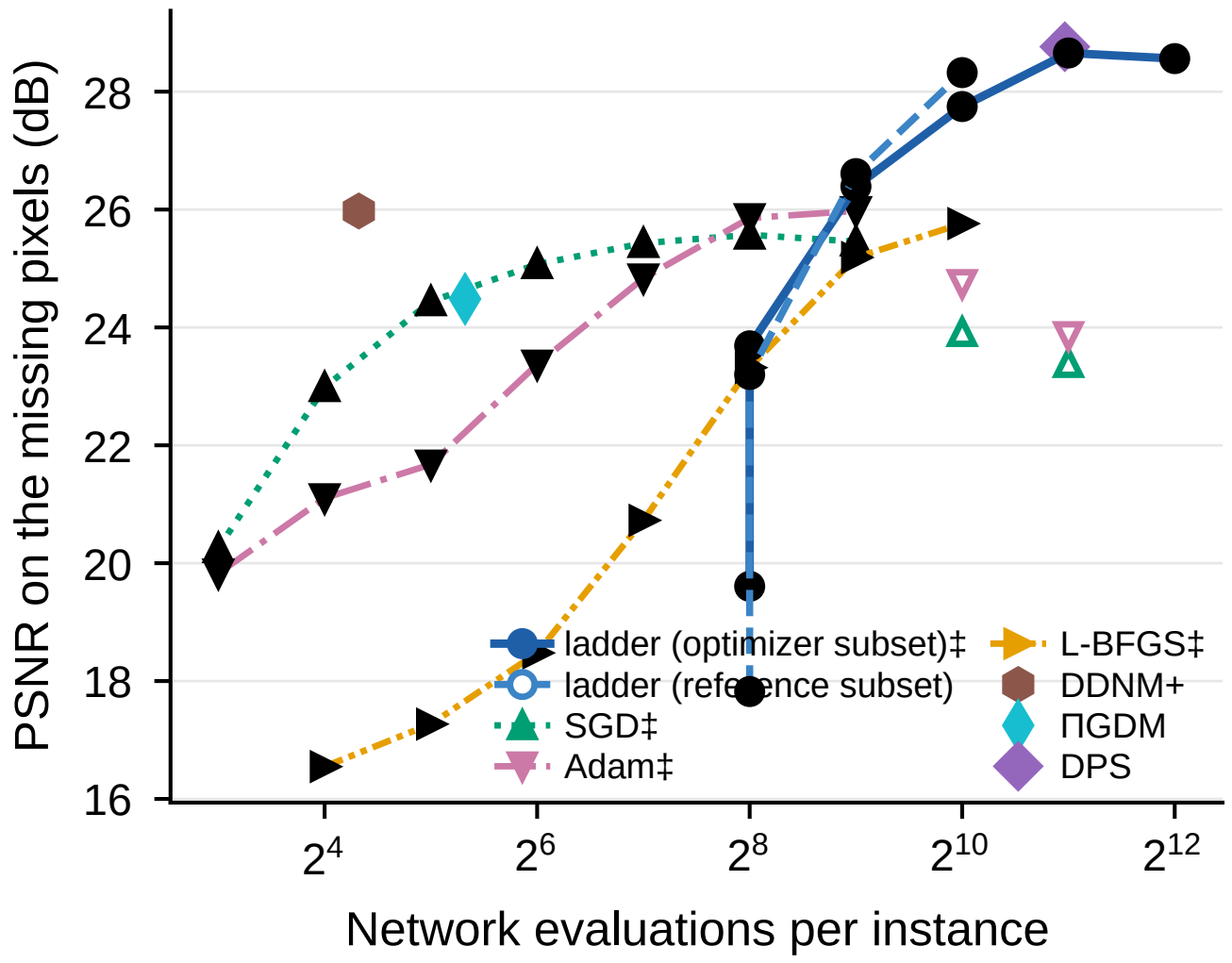}
\caption{Outcome against compute on one common denominator of network evaluations, on random-mask
inpainting. The ladder is drawn on both instance subsets because the optimizer runs used a disjoint set
($\ddag$). The specialized inverse methods buy early reconstruction gains at lower network cost.
Optimizer lines use the 50-image cohort; hollow optimizer markers are separate 20-image
extension endpoints. The paired reconstruction-versus-residual trajectory is in Fig.~\ref{fig:adam-collapse}.}
\label{fig:cost}
\end{figure}

Figure~\ref{fig:cost} puts reconstruction outcomes on a shared network-evaluation
axis and distinguishes the evaluated image populations. The accounting below
explains how the local inverse and optimizer updates contribute to that cost;
the solver summary specifies the corresponding implementations.

One Golub--Kahan iteration requires one VJP and one JVP through
the remaining sampler. With the 16-iteration cap, one matrix-free
correction uses up to 16 VJPs and 16 JVPs; $K$ corrections use
up to $16K$ of each. For the vision configuration with eight
remaining denoising steps, this corresponds to 256
derivative-traversal equivalents per correction, versus 128
for 16 SGD/Adam steps. These counts exclude primal forward
evaluations, checkpoint recomputation, and endpoint evaluation.
Reported timings on the shared GPU are approximately 6.55 s
per Krylov iteration and 3.44 s per SGD/Adam step.
Magnitude-matched FO reuses K1's computation in the directional
control; this shared accounting is not its standalone cost.

\subsection{Solver implementation}\label{app:solver-implementation}
The same local inverse is evaluated with different derivative representations
according to the output dimension and generator. The table maps the experiments
to explicit or matrix-free implementations; the paragraphs below specify damping,
numerical support, and the iterative stopping rule.

\begin{center}
\small
\begin{tabularx}{\linewidth}{@{}p{0.30\linewidth}X@{}}
\toprule
Experimental setting & Jacobian and local solve \\
\midrule
CelebA-HQ restoration; waveform localized EQ and bandwidth extension &
Matrix-free JVPs/VJPs; 16-step Golub--Kahan projection and a direct damped solve in the projected space. \\
CIFAR and SD1.5 structured edits; mel-audio &
Explicit reverse-mode Jacobian; reduced SVD and spectral filtering. \\
Executed robotics &
Explicit Jacobian with batched reverse-mode rows sharing a forward graph; row-wise fallback; reduced SVD. \\
Early robot geometry diagnostics &
Explicit sampler Jacobian composed with the objective Jacobian by the chain rule; reduced SVD. \\
Character motion generation &
Explicit Jacobian through the remaining sampler and joint reconstruction; batched reverse-mode rows with a row-wise fallback; reduced SVD (Appendix~\ref{app:motion}). \\
Custom trajectory control and standard 2D planning &
Explicit reverse-mode Jacobian and reduced SVD; 2D planning uses batched rows with a row-wise fallback. \\
\bottomrule
\end{tabularx}
\end{center}

\paragraph{Damping and support.}
In the audited implementations, ``adaptive'' means
$\lambda=0.01\widehat\sigma_1^2$, refreshed at every local correction.
Explicit solvers use $\widehat\sigma_1=\sigma_1(J)$ and apply
$V\operatorname{diag}[\sigma_i/(\sigma_i^2+\lambda)]U^\top d$ without
hard rank truncation. The SD1.5 reachability diagnostic retains
$\sigma_i>10^{-4}\sigma_1$; that campaign separately excludes
$\sigma_1<10^{-8}$ (an absolute threshold).
The common audio/control/robotics cores log numerical rank using
$\sigma_i>10^{-6}\sigma_1$; this is a separate diagnostic.
Named damping ablations vary the relative coefficient: the SD1.5 helper
defines $\gamma\in\{10^{-4},10^{-3},10^{-2},10^{-1},1\}$, and the
inpainting runs use $0.1\times$ and $10\times$ the default.
The SD1.5 fixed-damping control instead uses
$\lambda_0=0.01\operatorname{median}_{\rm pilot}(\sigma_1^2)$ per objective
and intervention time. None of these rules adds a damping line search.

\paragraph{Matrix-free solve.}
Starting from $u_1=d/\|d\|_2$, Golub--Kahan bidiagonalization constructs
$U,V,B$ with full reorthogonalization against the stored vectors of both bases.
The cap is 16 iterations; early termination occurs only if a newly computed
unnormalized basis vector has norm at most $10^{-30}\|d\|_2$.
There is no residual-based stopping tolerance or preconditioner.
Basis storage and projected algebra use float64; derivative evaluations retain
the model's dtype. The implementation sets
$\widehat\sigma_1=\sigma_1(B)$ using an SVD of $B$, then computes
\[
z=(B^\top B+\lambda I)^{-1}B^\top(\|d\|_2e_1),\qquad
\Delta x=Vz.
\]
No separate power iteration is used. The logged residual is that of these
\emph{projected} normal equations, not a stopping test or a full-space
convergence certificate. A zero request returns a zero correction.
The request-started projection is not used to establish full-Jacobian
numerical reachability.

\section{Universal Guidance Protocol}\label{app:ugprotocol}

We distinguish selected guidance settings from an explicit instability audit.
Table~\ref{tab:guidance} records failures at one unstable setting per vision
operator; it is not the selected-setting comparison in the main figure.
The remaining text specifies the vision, audio, and control-and-planning protocols.

\begin{table}[!htbp]
\centering\small
\caption{\textbf{Failures below the unguided prior at an unstable guidance setting.}
Counts for Universal Guidance (UG) use one evaluated unstable setting per
operator ($n=50$); spectral counts pool its three evaluated rungs ($n=150$).
These are robustness counts, not the post-hoc best-setting PSNR comparison
shown in Fig.~\ref{fig:vision-reconciled}.}
\label{tab:guidance}
\begin{tabular}{lrr}
\toprule
Vision operator & UG below prior & Spectral below prior\\
\midrule
Inpainting & 4/50 & \textbf{0/150}\\
$4\times$ super-resolution & 50/50 & \textbf{0/150}\\
Deblurring & 40/50 & \textbf{0/150}\\
Multi-constraint & 50/50 & \textbf{0/150}\\
\bottomrule
\end{tabular}
\end{table}
Figure~\ref{fig:vision-reconciled} reports Universal Guidance alongside the
vision baselines, using the selected settings from the frozen operator audit:
fixed $m=2$ for inpainting, fixed $m=1$ for super-resolution, and normalized
$m=8$ for deblurring and multi-constraint restoration.

On the 50 reference inpainting images, selected UG reaches 28.39 dB versus K4's 28.32, with no significant paired difference. The dense Adam trajectory peaks at 25.60 dB at 54 steps (Fig.~\ref{fig:adam-collapse}).

The separate audio guidance comparison uses the unconstrained $Z=16$ active-preservation setting.
For control and planning, the character motion generation adaptation is described
in Appendix~\ref{app:motion}. The constrained-planning results use the same-loop reruns with all methods paired
($n=315$ dense, $n=263$ novel obstacle); the separate UG comparisons use fixed $m=2$ and
fixed $m=16$, respectively. Figure~\ref{fig:rob:planning} uses these paired reruns.

The audio comparison uses forward guidance plus self-recurrence, with scale selected on separate tuning examples. At $Z=16$, unconstrained Universal Guidance reaches 0.0501 dB error with latent norm 58.8, versus K1's 0.4481 dB and norm 6.8; displacement matched, it reaches 1.5760 dB. At $Z=32$, the corresponding values are 0.0327 versus 0.5308 dB unconstrained and 1.5356 versus 0.5308 dB matched. This is evidence that nonlinear guidance can optimize the endpoint and that movement control changes the question.

\FloatBarrier
\newcommand{\qualrow}[2]{%
  \includegraphics[width=\textwidth,height=.205\textheight,keepaspectratio]{#1}\par
  \vspace{1pt}{\small #2}\par\vspace{3pt}}
\newcommand{\robotrow}[1]{\includegraphics[width=.97\textwidth,keepaspectratio]{#1}\par\vspace{3pt}}
\newcommand{\qualrowlarge}[2]{%
  \includegraphics[width=\textwidth,height=.30\textheight,keepaspectratio]{#1}\par
  \vspace{1pt}{\small #2}\par\vspace{4pt}}
\newcommand{\planningpanel}[3]{%
  \begin{subfigure}[t]{.485\textwidth}\vspace{0pt}\centering
  \begin{minipage}[c][#1][t]{\linewidth}\vspace{0pt}\centering
  \includegraphics[width=\linewidth,height=#1,keepaspectratio]{#2}%
  \end{minipage}\par\vspace{2pt}
  {\small #3}\par
  \end{subfigure}}
\newcommand{\gridpanel}[2]{\planningpanel{.205\textheight}{#1}{#2}}
\newcommand{\gridpanelsmall}[2]{\planningpanel{.21\textheight}{#1}{#2}}
\newcommand{\planningrow}[2]{%
  \noindent\makebox[\textwidth][c]{#1\hspace{.03\textwidth}#2}\par}

\section{Qualitative Results}\label{app:qualitative-gallery}
The gallery complements the examples alongside the domain plots in
Figs.~\ref{fig:method}, \ref{fig:vision-reconciled}, and~\ref{fig:rob}. Each panel is an existing experiment output.
The organizing question is whether the requested coordinates change while the named
preservation coordinates and downstream behavior remain satisfactory.

\paragraph{Selection and interpretation.}
Captions identify recorded median, quantile, boundary, and high-effect examples.
Video-derived control and planning panels use synchronized instants selected for
maximum change-from-initial separation between FO and the highest displayed
spectral rung. These are illustrations, not estimates of typical performance;
printed metrics describe individual cases. Crops preserve aspect ratios.
Task definitions and endpoints are in Appendix~\ref{sup:sec:task-objectives};
aggregate comparisons appear in the quantitative results.

\subsection{Vision}\label{sup:sec:qual-vision}

\begin{figure*}[!htbp]\centering
\qualrow{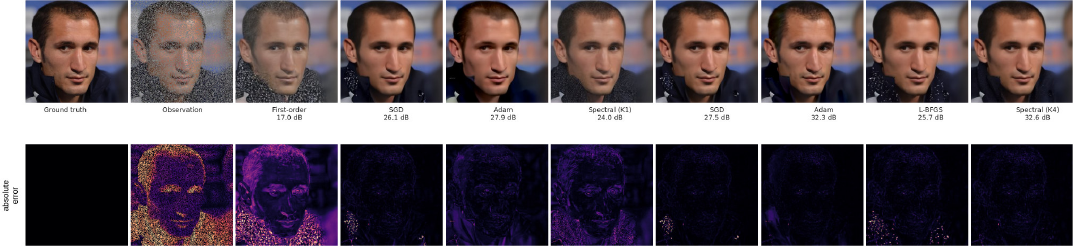}{Frozen optimizer-media case with SGD, Adam, and L-BFGS.}
\qualrow{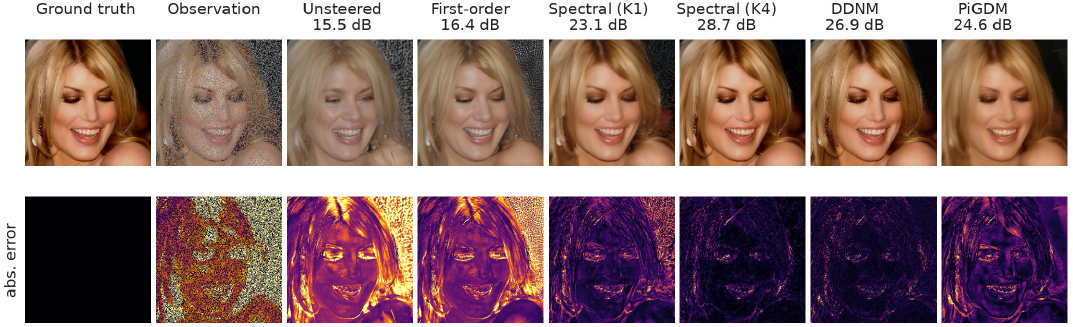}{Random-mask inpainting: source-frozen upper-quartile example.}
\qualrow{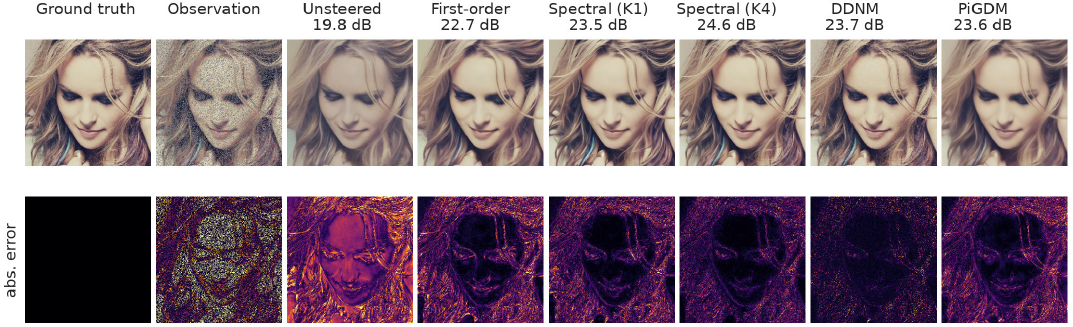}{Random-mask inpainting: source-declared boundary example.}
\caption{Random-mask inpainting: a frozen optimizer comparison, an upper-quartile case, and a boundary case. Methods share the observation within each example; printed PSNR and error maps describe that instance.}\label{sup:fig:qual-vision-inpaint}
\end{figure*}

\begin{figure*}[!htbp]\centering
\qualrow{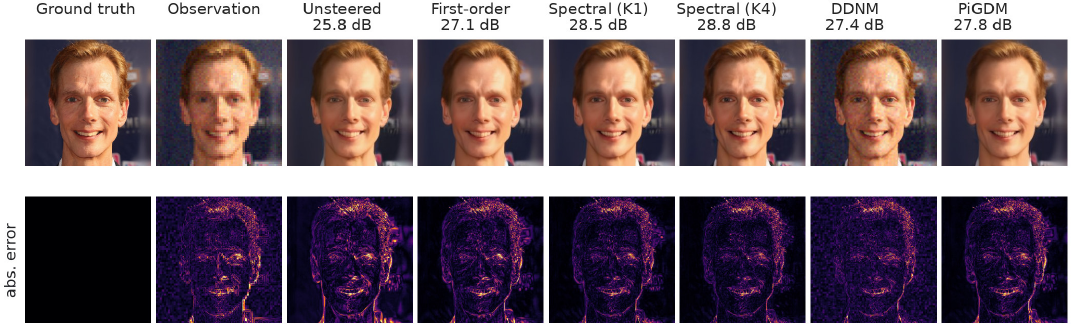}{$4\times$ super-resolution: source-frozen upper-quartile example.}
\qualrow{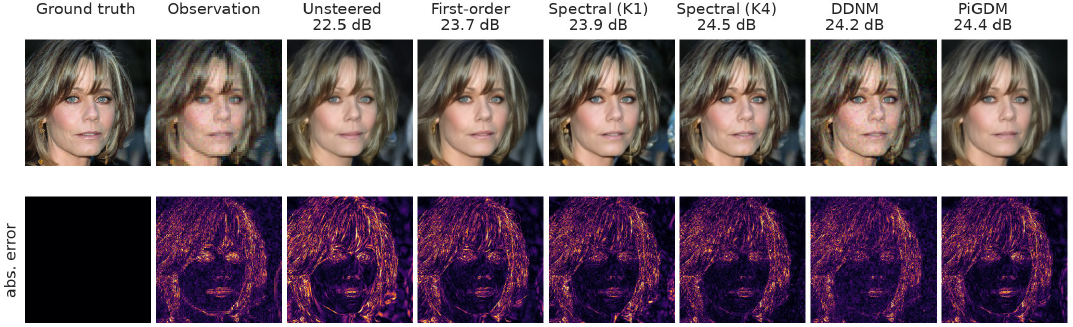}{$4\times$ super-resolution: source-declared boundary example.}
\qualrow{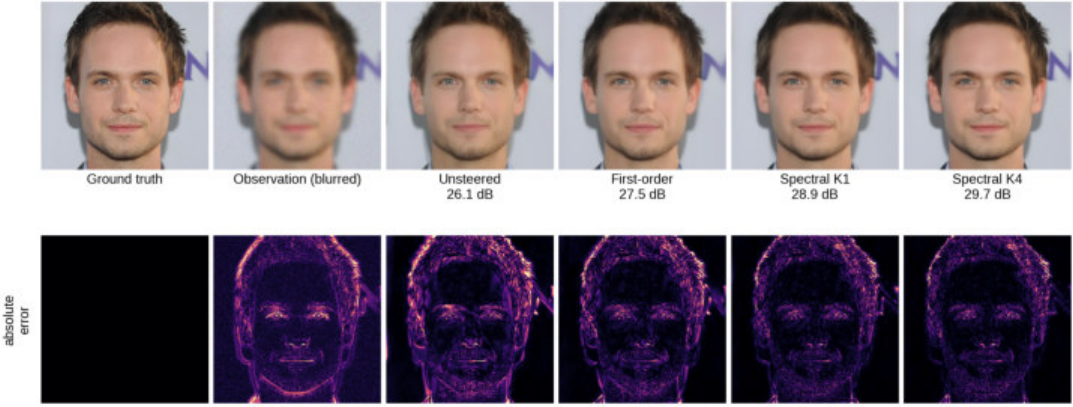}{Gaussian deblurring: source-frozen median example.}
\caption{Vision inverse-problem examples: super-resolution and deblurring.
The reconstruction and error maps complement PSNR by locating residual errors.
The upper-quartile, boundary, and median cases illustrate variation within the stored evaluations.}
\label{sup:fig:qual-vision-restoration}
\end{figure*}

\begin{figure*}[!htbp]\centering
\qualrowlarge{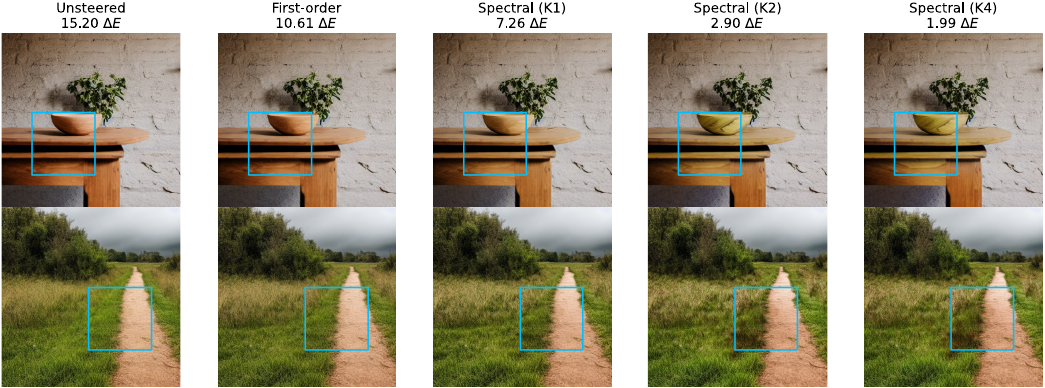}{Joint edit of two images: the outlined regions are requested to agree while the remaining cells are held fixed.}
\qualrow{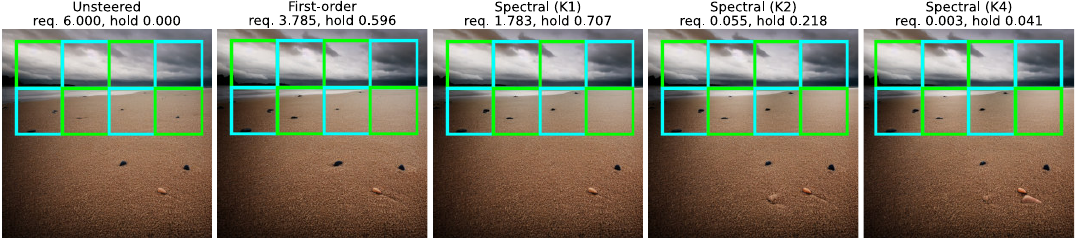}{Multi-region appearance control with requested and held CIELab blocks.}
\caption{Paired-image and multi-region objectives from the vision extension.}
\label{sup:fig:qual-vision-extended}
\end{figure*}

Figures~\ref{sup:fig:qual-vision-inpaint}--\ref{sup:fig:qual-vision-restoration} show inverse problems whose
loss is a measurement residual. Figure~\ref{sup:fig:qual-vision-extended} shows paired-image and multi-region
objectives built from regional color, agreement, and preservation coordinates. Printed metrics describe
the displayed instances.

\FloatBarrier
\subsection{Audio}\label{sup:sec:qual-audio}

\begin{figure*}[!htbp]\centering
\qualrowlarge{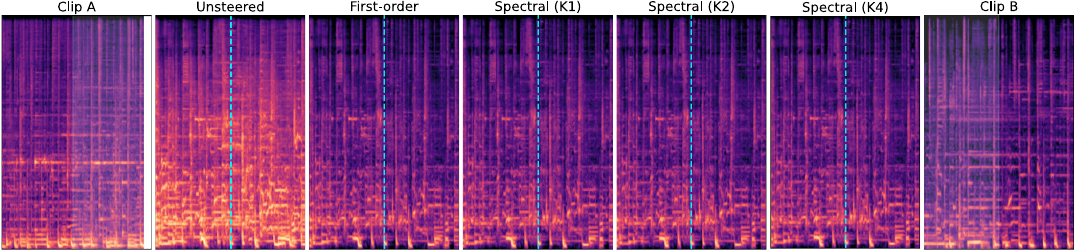}{Crossfade from clip A to clip B across four temporal quarters.}
\qualrowlarge{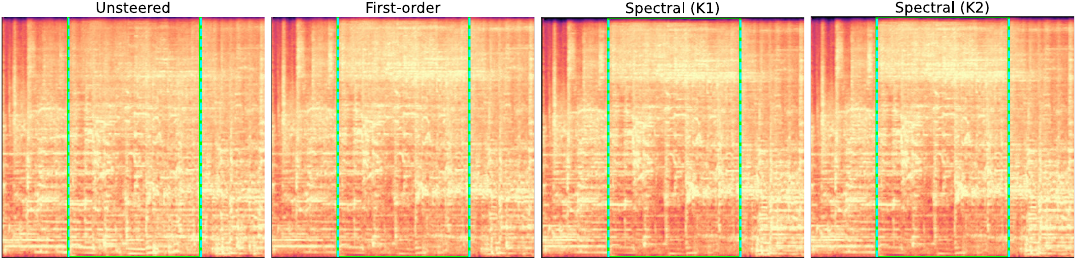}{Interior high-band boost and low-band cut with both contexts and both seams held.}
\caption{Audio transition objectives.  Dashed or boxed boundaries in the spectrograms mark the temporal regions entering the cost.}
\label{sup:fig:qual-audio-transitions}
\end{figure*}

Figures~\ref{sup:fig:qual-audio-eq}, \ref{sup:fig:qual-audio-restoration}, and~\ref{sup:fig:qual-audio-transitions} organize the audio examples by objective support.  Equalization rows expose the time--frequency
cells that enter  \eqref{sup:eq:audio-cell}; restoration rows expose observed and missing frequency or
time regions; transition rows expose the temporal quarters, contexts, and seams used by their dB
objectives.  Spectrograms locate changes in time and frequency. They are visual diagnostics; perceived quality requires listening evidence, and held-cell errors remain part of the quantitative evaluation.

\begin{figure*}[!htbp]\centering
\qualrow{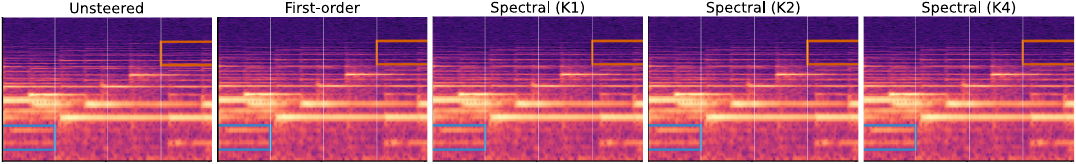}{Localized time--frequency equalization shown as spectrograms.}
\qualrow{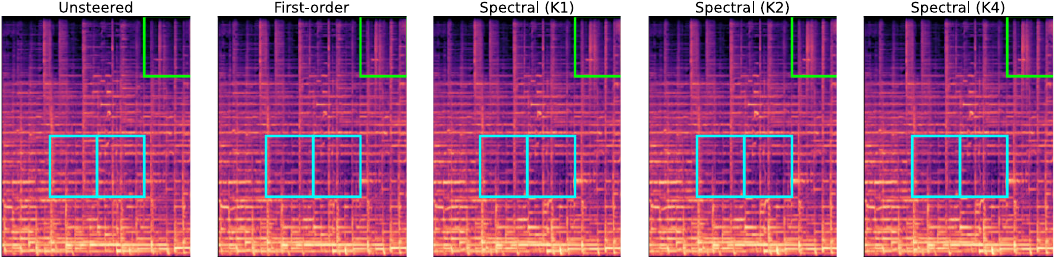}{Localized EQ automation: low-mid cut in the middle and high-band boost late in the clip.}
\caption{Localized and automated audio equalization. Requests identify both edited and held time--frequency regions; Appendix~\ref{app:audio-detail} reports their quantitative errors.}
\label{sup:fig:qual-audio-eq}
\end{figure*}

\begin{figure*}[!htbp]\ContinuedFloat\centering
\qualrow{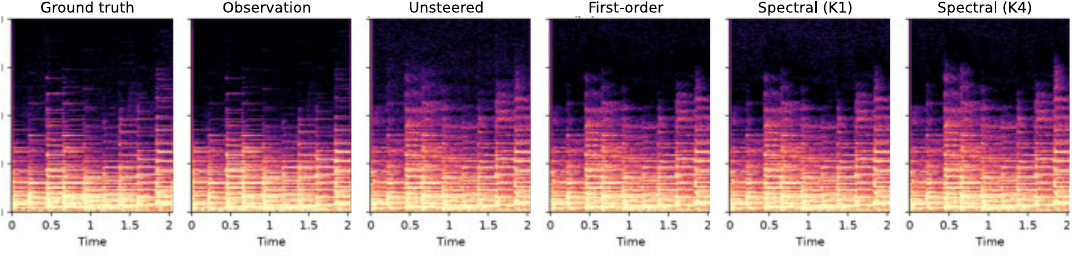}{Known-filter equalization restoration.}
\caption{Audio equalization (continued): restoration toward a supplied target spectrum.}
\end{figure*}

\begin{figure*}[!htbp]\centering
\qualrow{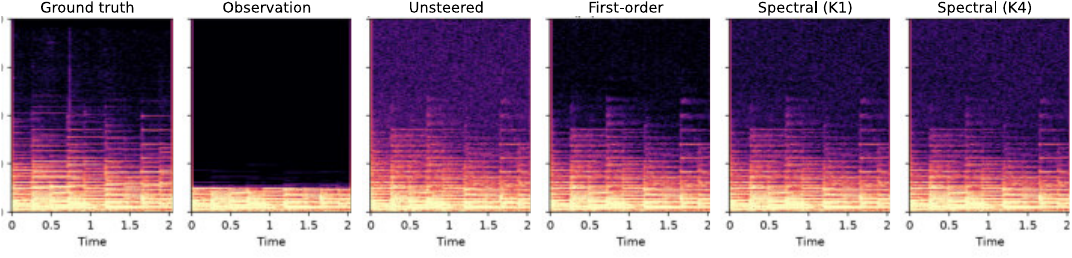}{Bandwidth-extension reconstruction.}
\qualrow{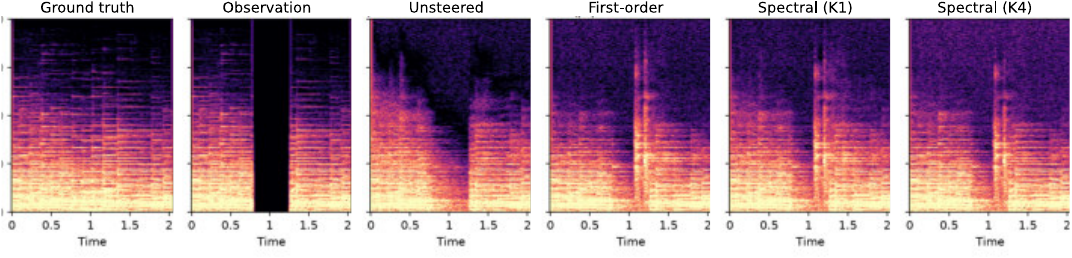}{Missing-segment waveform inpainting.}
\caption{Audio restoration: bandwidth extension and missing-segment inpainting.}
\label{sup:fig:qual-audio-restoration}
\end{figure*}

\begin{figure*}[!htbp]\ContinuedFloat\centering
\qualrow{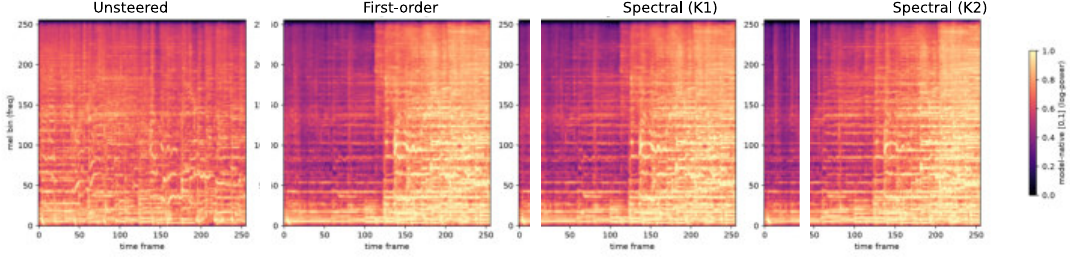}{Crescendo temporal-envelope edit shown as spectrograms.}
\caption{Audio restoration and temporal control (continued): crescendo-envelope editing.}
\end{figure*}

\FloatBarrier
\subsection{Robotics}\label{sup:sec:qual-robotics}

\begin{figure*}[!htbp]\centering
\robotrow{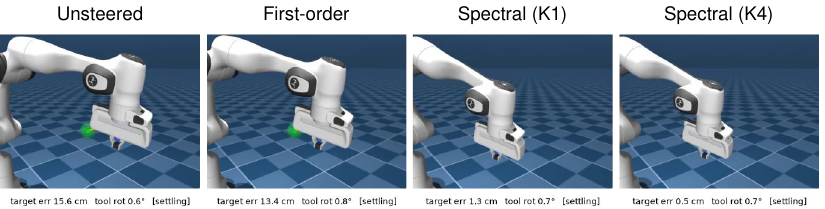}
\robotrow{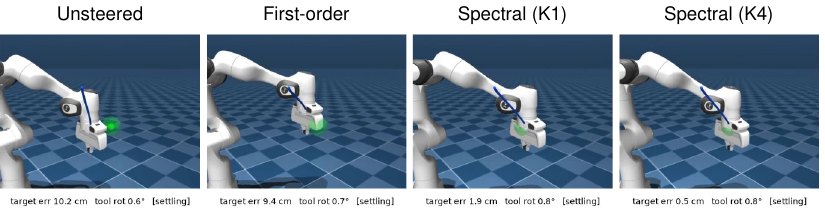}
\robotrow{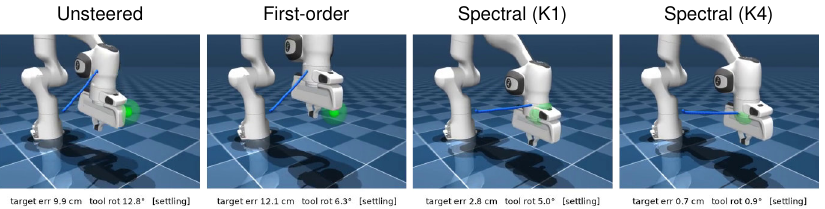}
\caption{Manipulation, top to bottom: Cartesian reach, XYZ-marker reach, and orientation hold.}
\label{sup:fig:qual-robotics-reach}
\end{figure*}

\begin{figure*}[!htbp]\centering
\robotrow{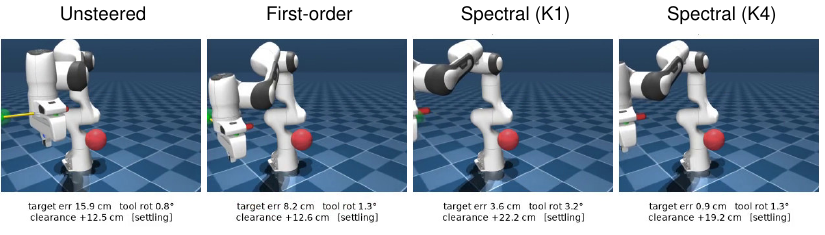}
\robotrow{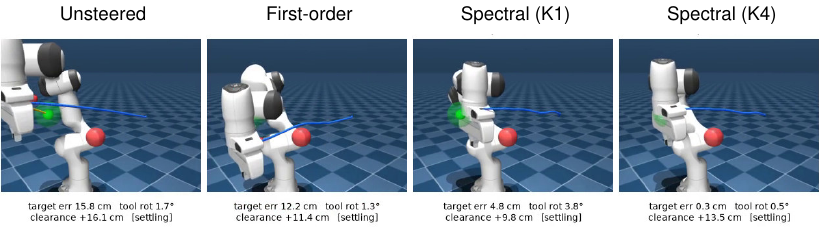}
\robotrow{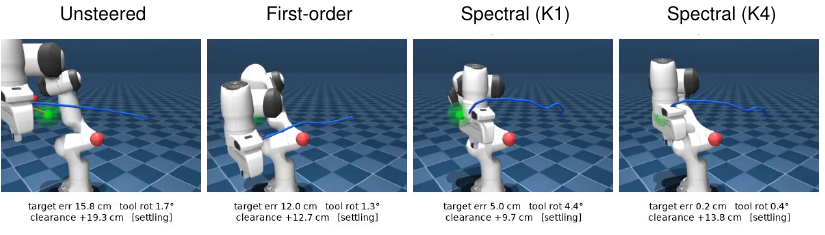}
\caption{Executed obstacle avoidance, top to bottom: unseen obstacle, lateral clearance, and narrow gap. Selected high-effect cases; contact-free execution must also complete the reach.}
\label{sup:fig:qual-robotics-obstacles}
\end{figure*}

Each row is a synchronized frame from the stored execution videos.  The first two groups show
position, orientation, checkpoint, and plane-preservation objectives (Figs.~\ref{sup:fig:qual-robotics-reach} and~\ref{sup:fig:qual-robotics-constraints}).  The third shows whole-arm
clearance constraints, including cases where the first-order rollout contacts the obstacle while a
spectral rollout remains clear (Fig.~\ref{sup:fig:qual-robotics-obstacles}). These manipulation examples were selected as high-effect illustrations.
A contact-free unsteered rollout that fails to complete the reach does not establish task success.

\begin{figure*}[!htbp]\centering
\robotrow{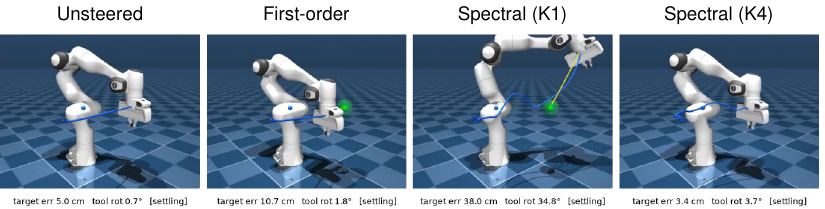}
\robotrow{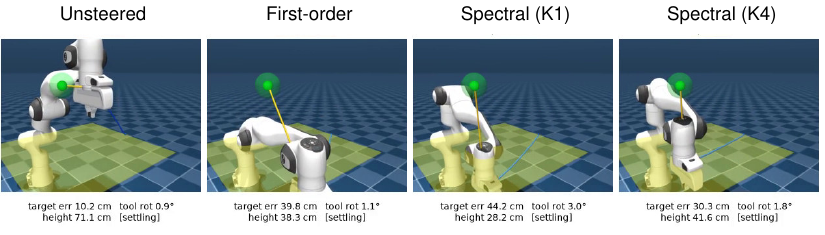}
\caption{Manipulation, top to bottom: via-point reach and constant-height plane constraint.}
\label{sup:fig:qual-robotics-constraints}
\end{figure*}

\begin{figure*}[!htbp]\ContinuedFloat\centering
\robotrow{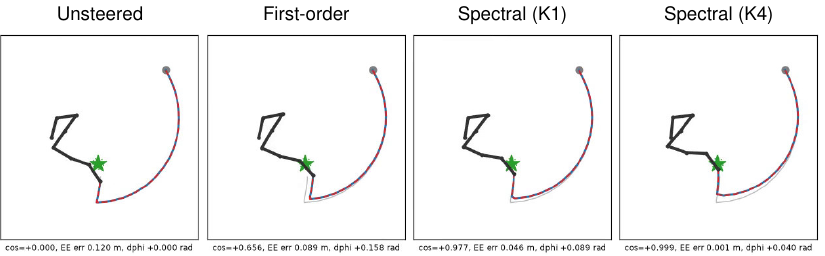}
\caption{Manipulation (continued): planar-arm pose control.}
\end{figure*}

\FloatBarrier
\subsubsection{Locomotion}\label{sup:sec:qual-locomotion}

\begin{figure*}[!htbp]\centering
\includegraphics[width=.96\textwidth,height=.68\textheight,keepaspectratio]{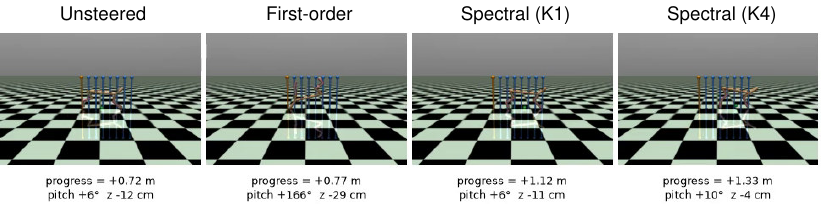}
\caption{Forward progress over eight planning windows.  All references are executed by the same objective-blind tracking controller.}
\label{sup:fig:qual-locomotion}
\end{figure*}

Figure~\ref{sup:fig:qual-locomotion} retains the rollout renderings and their printed progress and posture outcomes.  The
cost requests forward progress over eight replanning windows while holding torso height and pitch.
This source-declared high-effect example illustrates closed-loop transmission; the quantitative
reference-to-execution limitations remain in Appendix~\ref{app:boundaries}.

\FloatBarrier
\subsection{Control and planning}\label{sup:sec:qual-control-planning}
The gallery starts with character motion generation, where the specification
coordinates several joints over time. Canonical control and trajectory planning
follow, showing how the same interface applies to dynamical states and routes.

\subsubsection{Character motion generation}\label{app:motion-qualitative}
\begin{figure}[!htbp]
\centering
\includegraphics[width=.86\linewidth]{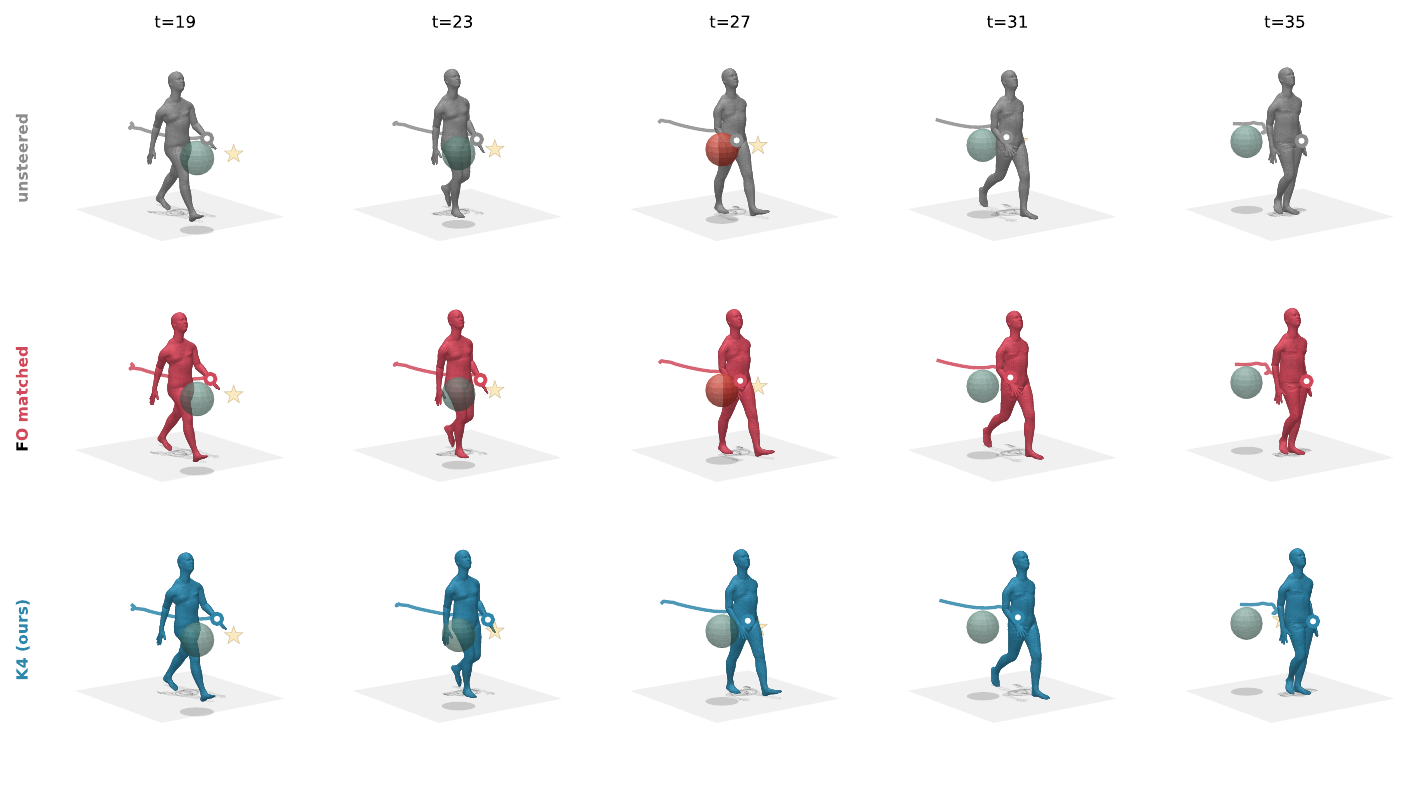}
\caption{\textbf{Obstacle-avoidance character sequence.} Five instants surround the
unsteered motion's deepest penetration. Rows show unsteered, magnitude-matched
first-order, and Spectral (K4). A red sphere indicates that any skeleton segment
intersects it. Figure~\ref{fig:method} uses two instants from this same sequence.}
\label{fig:motion-obstacle}
\end{figure}

\begin{figure}[!htbp]
\centering
\includegraphics[width=.86\linewidth]{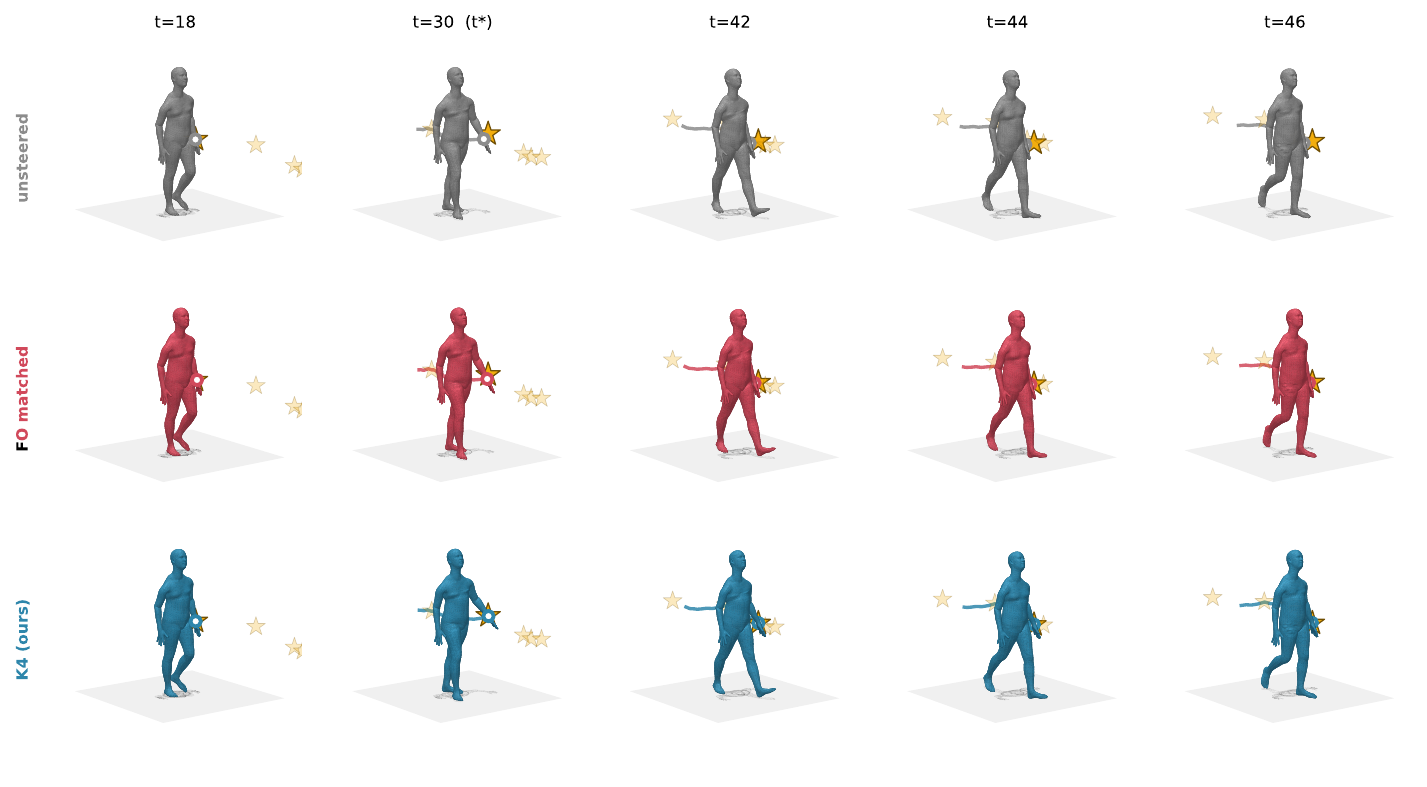}
\caption{\textbf{Multi-objective character reaching.} Columns show five requested
instants; rows show unsteered, magnitude-matched first-order, and Spectral (K4),
with a fixed camera. Stars mark requested wrist positions; the active waypoint is
solid. Foot positions, zero arrival velocity, and bone lengths enter the objective;
velocity and bone drift are separate diagnostics from the reported success criterion. The inherited small sphere is not a clearance objective for this task.}
\label{fig:motion-multi}
\end{figure}

Figures~\ref{fig:motion-obstacle} and~\ref{fig:motion-multi} show avoidance and reaching with the same camera and instance across methods.

\FloatBarrier
\subsubsection{Canonical control}\label{sup:sec:qual-control}

\begin{figure*}[!htbp]\centering
\includegraphics[width=.94\textwidth,height=.44\textheight,keepaspectratio]{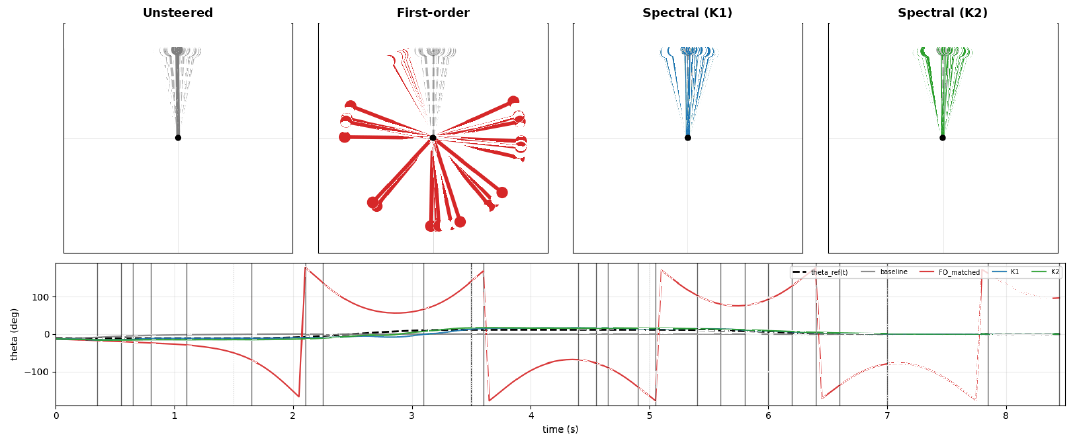}
\caption{Pendulum reference tracking, shown on a source-declared high-effect example. The angular reference and full traces complement the selected rendering.}
\label{sup:fig:qual-control}
\end{figure*}

\begin{figure*}[!htbp]\centering
\qualrow{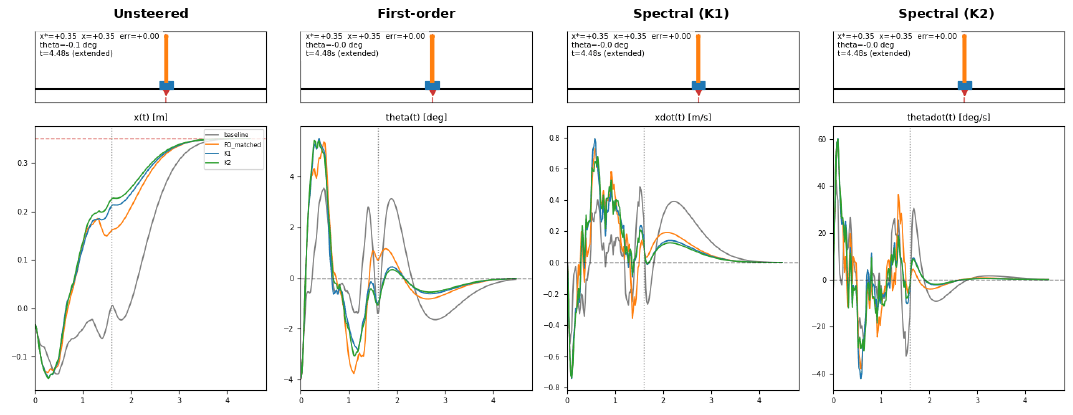}{CartPole terminal-position request: selected illustrative rollout.}
\caption{CartPole terminal displacement with upright-angle and zero-velocity terminal coordinates.
The upper strip renders the selected state, while the lower plots show the full state trajectories and
the requested terminal cart position.}
\label{sup:fig:qual-cartpole}
\end{figure*}

The pendulum request (Fig.~\ref{sup:fig:qual-control}) specifies checkpoint angles and terminal angular velocity;
CartPole requests a positive terminal cart displacement with upright, stationary terminal targets (Fig.~\ref{sup:fig:qual-cartpole}).
The renderings are paired with full state traces, so a selected frame can be checked against
the complete rollout. Frame selection follows the visual-separation rule above.

\FloatBarrier
\subsubsection{Trajectory planning}\label{sup:sec:qual-planning}

\begin{figure*}[!htbp]\centering
\planningrow{\gridpanelsmall{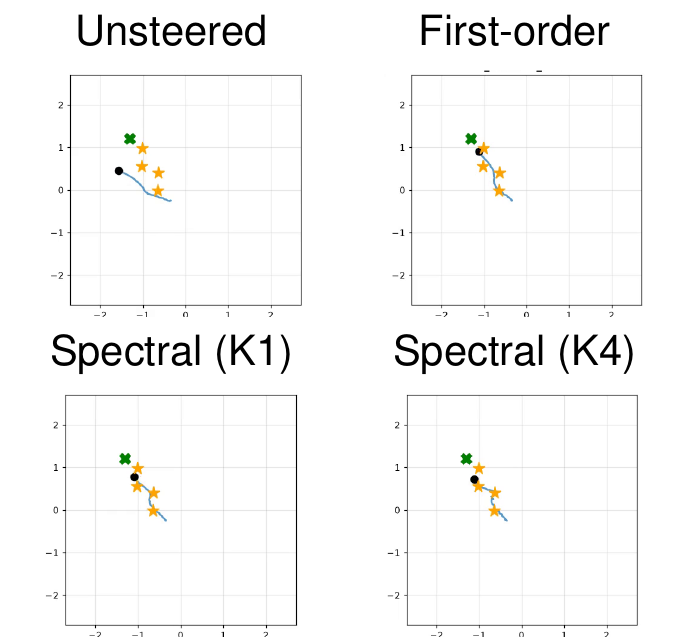}{Four timed waypoints.}}{%
\gridpanelsmall{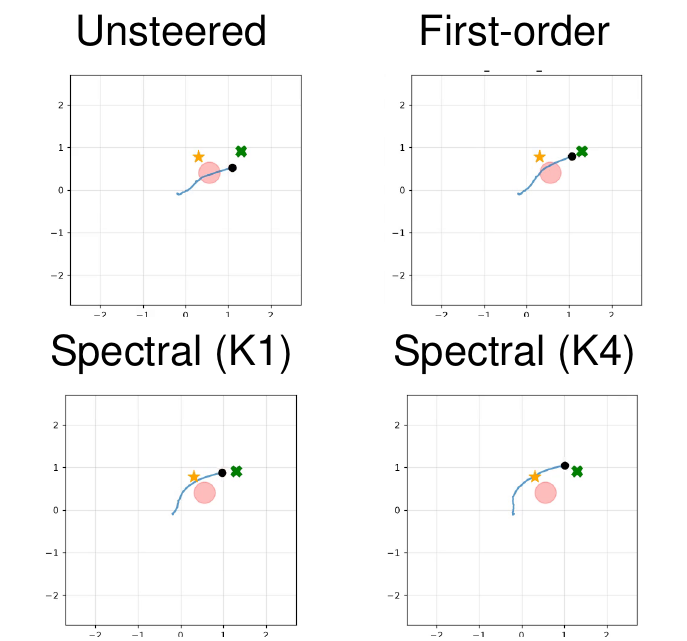}{Waypoint and goal with obstacle avoidance.}}
\caption{Double-integrator trajectory objectives.  Method order is fixed as Unsteered,
First-order, Spectral (K1), and Spectral (K4).}
\label{sup:fig:qual-double-integrator}
\end{figure*}

Figure~\ref{sup:fig:qual-double-integrator} shows timed-waypoint tracking and obstacle avoidance with a double integrator. Black dots mark endpoints, orange stars waypoints, green
crosses goals, and red disks obstacles. Frames maximize the change-from-initial
separation between FO and Spectral (K4), following the gallery selection rule.

\begin{figure*}[!htbp]\centering
\planningrow{\gridpanel{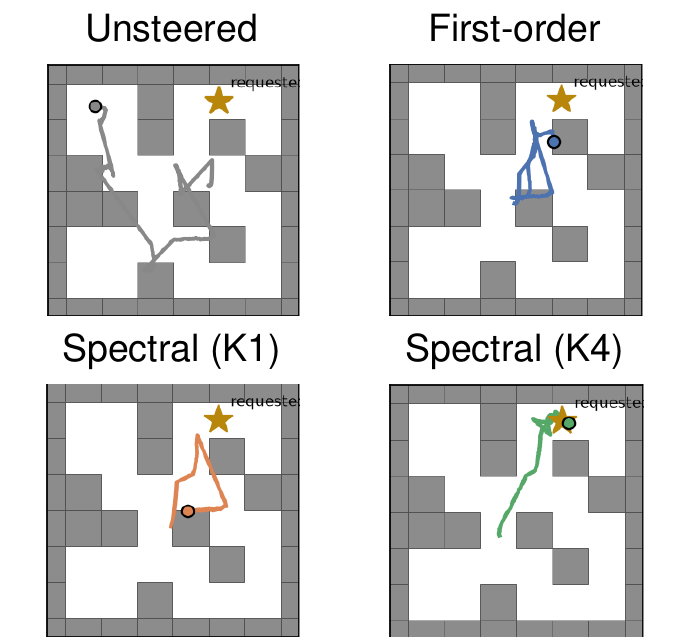}{Maze2D medium maze, shifted goal.}}{%
\gridpanel{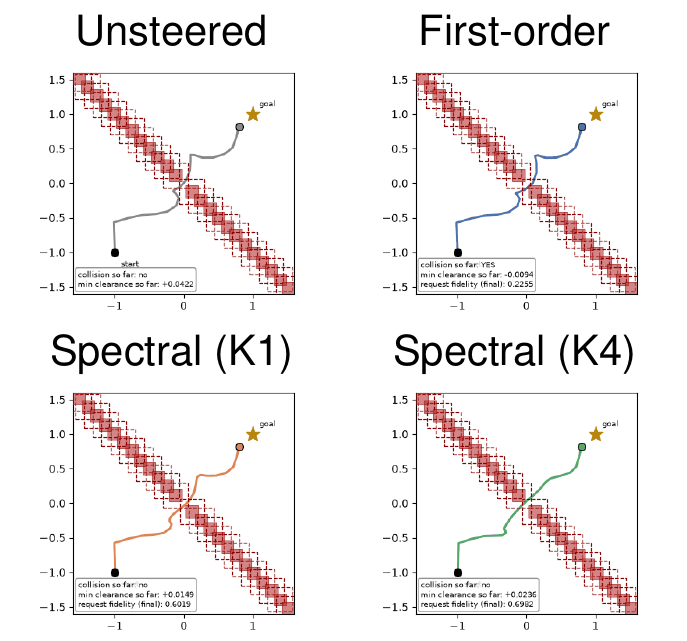}{Narrow-passage planning.}}
\caption{Goal-conditioned and obstacle-aware diffusion planning.  Each trajectory panel uses the same
method order: Unsteered, First-order, Spectral (K1), and Spectral (K4).}
\label{sup:fig:qual-planning-a}
\end{figure*}

Figure~\ref{sup:fig:qual-planning-a} shows a shifted Maze2D goal and a narrow passage.
Path diagnostics report goal error, clearance, and smoothness; negative
clearance denotes a violation. Paths are generated references unless
execution is explicitly reported.

\FloatBarrier
\section{Task Objectives and Evaluation Endpoints}\label{sup:sec:task-objectives}
The following catalogue connects the common steering interface to the task-specific
experiments and preceding gallery. It supplements the quantitative comparisons with the definition
of what each method is asked to change, what is explicitly held, and how the outcome is evaluated.
Parenthesized source keys identify the corresponding task campaigns.

\paragraph{Datasets and frozen diffusion models.}\label{app:diffusion-models}
The model names in Table~\ref{tab:cross-domain} abbreviate the actual frozen
checkpoints. Vision restoration in the main table uses \path{google/ddpm-ema-celebahq-256} (DDPM EMA-256)~\citep{ho2020denoising}
on CelebA-HQ validation images~\citep{xia2021tedigan}. The structured vision edits use an additional model:
\path{stable-diffusion-v1-5/stable-diffusion-v1-5} (Stable Diffusion 1.5)~\citep{rombach2022high},
with generated 512$\times$512 images from the campaign's frozen prompt/seed pools.
This is the model in Fig.~\ref{fig:vision-reconciled}(e), the spatial and composite
editing experiments, and the paired-image/region-transfer qualitative examples.
The table uses standard-scale edits; the guidance figure uses $4\times$ requests.

Waveform EQ and bandwidth extension use
\path{harmonai/maestro-150k} (MAESTRO 150k) on MAESTRO v3 test excerpts~\citep{hawthorne2018enabling}.
EQ-16 and selective EQ use \path{teticio/audio-diffusion-ddim-256}
(Audio Diffusion), evaluated on generated log-mel spectrograms.

Temporal DDPM denotes a task-specific Diffuser-style temporal U-Net, trained
before steering and then frozen. Panda reaching uses \path{m1_generator_v1.pt};
Panda pose and obstacle-aware reaching use \path{m2_generator_v1.pt}.
Both HalfCheetah rows use \path{loco_halfcheetah_generator_v2.pt}.
Execution uses MuJoCo~\citep{6386109}: Panda position servos or a custom
tracking wrapper around Gymnasium's HalfCheetah model~\citep{robomimic2021}.
The character extension uses a frozen text-conditioned MDM~\citep{tevet2023human}
(60 frames at 20 fps, 50 DDIM steps, classifier-free guidance 2.5), with
25 deterministic transitions remaining at the intervention, acting on the
$263\times1\times60$ HumanML3D feature representation.
We use the HumanML3D MDM checkpoint
\texttt{humanml\_enc\_512\_50steps/model000750000.pt},
with a 512-dimensional transformer encoder and eight layers.

The custom two-dimensional double-integrator experiments use distinct frozen
checkpoints: \path{di_generator.pt} for one-waypoint control,
\path{ds_generator.pt} for multi-waypoint tracking, and
\path{r1_generator_v5.pt} for the point-mass cosine task.
Sharing an architecture label does not imply sharing weights across these tasks.

\subsection{Request, preservation, and outcome}\label{sup:sec:common}
For the frozen remaining sampler in \eqref{eq:full-jacobian}, write
$z_0=h(\Frem(x_t^{(0)};c))$ and specify an absolute target $z_\star=z_0+d$.
The nonlinear request loss optimized by the generic gradient comparators is
\begin{equation}
\mathcal L_{\rm req}(x_t)=\tfrac12\|h(\Frem(x_t;c))-z_\star\|_2^2.
\label{sup:eq:nonlinear-loss}
\end{equation}
Coordinates use the units and any fixed scaling declared by the task; a pixel residual,
a dB error, and a normalized physical-state error are not interchangeable quantities.
A zero displacement $d_i=0$ holds a measured coordinate at its unsteered value.
An absolute target such as terminal velocity zero instead specifies
$d_i=-z_{0,i}$, unless the anchor is already stationary. Coordinates absent from $h$
receive no preservation constraint.

Spectral steering solves the damped local model in  \eqref{eq:local-ls}--\eqref{eq:woodbury};
K denotes the number of local solves at the same diffusion time. Generic SGD, Adam,
and L-BFGS optimize the nonlinear loss through the remaining sampler, adding a
fixed displacement penalty where the source protocol specifies one
(as in Appendix~\ref{app:one-waypoint}), with the
resource conventions in Appendices~\ref{app:adam-budget} and~\ref{app:optimizers}.
The magnitude-matched FO control isolates direction; the external guidance methods
retain the operator-specific protocols and disclosed adaptations of Fig.~\ref{fig:vision-reconciled}.

Request fidelity measures how well $\Delta h$ realizes $d$, for example
$E_{\rm req}=\|\Delta h-d\|_2/\|d\|_2$ for nonzero $d$.
PSNR on unobserved pixels and physical execution provide separate tests of whether this
realization is useful. The catalogue labels fidelity diagnostics separately from independent
endpoints; neither an accurate request nor a convincing single example establishes a
population-level quality or execution advantage.

\subsection{Vision}

The inverse-problem objectives compare a generated image after a known measurement operator with the
stored observation.  Structured edits instead reduce the image to analytic regional statistics.
Background or boundary preservation is part of the loss only when those coordinates are explicitly
included with zero requested change.  The vision gallery mirrors these two groups: restoration
strips first, then regional, paired-image, and multi-region edits. Table~\ref{sup:tab:vision-objectives} specifies their measurement maps and endpoints.

\begin{table*}[!htbp]
\centering\small
\caption{Vision objective and cost functions.  Pixel-space inverse problems use the measurement residual itself; structured editing uses analytic color, spatial, boundary, and texture statistics.}
\label{sup:tab:vision-objectives}
\setlength{\tabcolsep}{3pt}
\begin{tabularx}{\textwidth}{@{}>{\raggedright\arraybackslash}p{.16\textwidth}Y>{\raggedright\arraybackslash}p{.21\textwidth}@{}}
\toprule
Task / source & Measurement map and target & Evaluation \\
\midrule
Random-mask inpainting (V0) & Retained pixels $M\odot y$ in DDPM data units. \emph{Target:} $d=y_{\rm obs}-h(y_0)$; unobserved pixels are omitted, not preservation coordinates & PSNR on unobserved pixels \\
$4\times$ super-resolution (V0) & DPS bicubic antialiased resizer $R_{1/4}(y)$. \emph{Target:} $d=y_{\rm obs}-R_{1/4}(y_0)$ & Full-image PSNR \\
Gaussian deblurring (V0) & Reflection-padded $61\times61$ Gaussian convolution $G_{\sigma=3}y$. \emph{Target:} $d=y_{\rm obs}-G_{\sigma=3}y_0$ & Full-image PSNR \\
Two-region restoration (V0) & $[\operatorname{avgpool}_4(y|_{R_1});\ y|_{R_2}]$. \emph{Target:} Restore low-resolution $R_1$; preserve exactly observed $R_2$ & PSNR on $R_1$ \\
Multi-region color control (V0) & Mean CIELab coordinates in each named region, $Z\in\{12,27,48\}$. \emph{Target:} Fixed per-region color displacement; included preservation regions request zero change & Dense $\Delta E$ on preserved pixels \\
Spatial-statistic editing (V1) & $[\mu_x,\mu_y,\sigma_x,\sigma_y]$ under the smooth luminance weight $\widetilde w_{ij}$. \emph{Target:} Shift the weighted center and spreads; $d$ is calibrated from frozen pilot statistics & Objective fidelity \\
Combined color--geometry--texture (V1) & $[\bar R_B,\bar G_B,\bar B_B,\mu_x,\mu_y,E_{\rm HP}]$, where $E_{\rm HP}=\|K_{\rm Lap}*Y\|_2^2/N$. \emph{Target:} Pilot-scaled displacement in the raw objective coordinates; the apple large-edit illustration uses this objective at $4\times$ scale & Objective fidelity and outside-region change \\
Paired compositing/appearance family (V2) & Analytic CIELab destination cells, luminance spread, inner/outer boundary rings, and background preservation blocks. \emph{Target:} Reference-derived requested blocks; background entries are explicitly zero & Boundary and appearance diagnostics not optimized as semantic scores \\
Reference patch transfer (V3) & CIELab statistics over 20 destination edit coordinates and 56 background coordinates. \emph{Target:} Match the supplied source patch in the destination rectangle; request zero change outside it & Dense outside-region $\Delta E$ \\
Two-image joint edit (V3) & Concatenated regional CIELab cells for two jointly steered images, followed by preserve cells from both images. \emph{Target:} Drive the two named regions to agreement; request zero change over the remaining cells & Region disagreement and per-image collateral change \\
\bottomrule
\end{tabularx}
\end{table*}

The spatial-moment map uses normalized smooth luminance weights
$\widetilde w_{ij}\propto\operatorname{softplus}(4(Y_{ij}-\bar Y))+10^{-6}$
on an image grid in $[-1,1]^2$. In the combined objective, pilot standard deviations
scale the requested direction before its normalization in the raw measurement space;
they do not imply that every loss coordinate was whitened.
The separate benchmark in Fig.~\ref{fig:vision-reconciled}(e) uses the linear RGB
box-average map, at diffusion window 10 and request scale $4\times$ (source V4).
It is distinct from the combined color--geometry--texture apple illustration.

\subsection{Audio}

For waveform localized equalization, with STFT $X$, four temporal segments and six log-frequency
bands,
\begin{equation}
 h(x)_{q,b}=10\log_{10}\!\left(\sum_{(m,f)\in(q,b)}|X_{m,f}|^2+10^{-10}\right).
 \label{sup:eq:audio-cell}
\end{equation}
The other audio rows in Table~\ref{sup:tab:audio-objectives} either use the same dB construction on a different partition or use the stated
linear measurement operator directly.  For a dB-cell task, the target is additive: requesting
$+6$ dB means that the target coordinate is the unsteered value plus six.  Cells named as ``held''
are included with a zero displacement, so leakage into them increases the same request loss.  The
spectrogram gallery shows where those requested and held regions lie in time and frequency.

\begin{table*}[!htbp]
\centering\small
\caption{Audio objective and cost functions.  All dB requests are literal changes in the stated time--frequency cells.  No listening-study or perceptual-preference endpoint is claimed.}
\label{sup:tab:audio-objectives}
\setlength{\tabcolsep}{3pt}
\begin{tabularx}{\textwidth}{@{}>{\raggedright\arraybackslash}p{.16\textwidth}Y>{\raggedright\arraybackslash}p{.21\textwidth}@{}}
\toprule
Task / source & Measurement map and target & Evaluation \\
\midrule
Localized waveform EQ (A0) &  \eqref{sup:eq:audio-cell}, $4\times6=24$ dB cells. \emph{Target:} Primary request: $+6$ dB in one cell and $-8$ dB in another; remaining 22 cells request $0$ dB & Requested-cell dB error; LSD and SI-SDR to a reference edit \\
Disjoint-cell EQ ladder (A0) & Per-cell dB on $Z\in\{2,4,8,16,32\}$ disjoint log-mel cells. \emph{Target:} Alternating $\pm3$ dB pattern at the declared scales; included hold cells request $0$ change & Decoded-signal cell error and feasibility gate \\
Bandwidth extension (A1) & $A_{\rm LPF}(x)$, a 601-tap Kaiser low-pass measurement. \emph{Target:} $d=y_{\rm obs}-A_{\rm LPF}(x_0)$ & High-band LSD above the cutoff \\
Temporal inpainting (A1) & Observed samples of a centered binary temporal mask. \emph{Target:} Match the observed samples; the missing interval is absent from $h$ & Gap-region SNR and splice-boundary discontinuity \\
Known-filter EQ restoration (A1) & $24$ log-spaced band energies $10\log_{10}(\sum_{f\in b}|X_f|^2+10^{-10})$. \emph{Target:} $d=h(x_\star)-h(x_0)$, restoring the clean band balance & LTAS distance and waveform metrics \\
Missing-segment boundary match (A0) & Eight boundary-band levels in dB. \emph{Target:} Match the generated segment's beginning/end to the adjacent clips; interior coordinates request $0$ & Abrupt energy jump across the splice \\
Crossover between two clips (A3) & Four band levels in each of four temporal quarters, plus four broadband quarter levels ($Z=20$). \emph{Target:} Ramp each band and broadband level from clip A's ending profile to clip B's opening profile & Bandwise ramp error and splice diagnostics \\
Context-preserving interior edit (A4) & Four interior band levels; four left/right context levels; two seam-continuity coordinates ($Z=10$). \emph{Target:} Interior low band $-6$ dB and high band $+6$ dB; both contexts and both seams request no change & Interior error and leakage into contexts/seams \\
Crescendo envelope (A2) & $[E_1,\ldots,E_5,E_{\rm total}]$, five equal time-bin mean-dB levels and global mean dB. \emph{Target:} $8[-2,-1,0,1,2,0]$ dB; middle and total energy request zero change & Realized temporal envelope; no listening preference \\
\bottomrule
\end{tabularx}
\end{table*}

\subsection{Executed manipulation and locomotion}

The steering loss is applied to a generated reference trajectory, not directly to the MuJoCo rollout.
The same objective-blind controller then executes every method's reference.  This separation is why
Table~\ref{sup:tab:robot-objectives} distinguishes requested reference coordinates from executed target error, contact, height
drift, or world-frame progress.

\begin{table*}[!htbp]
\centering\small
\caption{Manipulation and locomotion objectives.  The objective is evaluated on the generated reference; the endpoint is measured after an independent controller executes that reference in MuJoCo.}
\label{sup:tab:robot-objectives}
\setlength{\tabcolsep}{3pt}
\begin{tabularx}{\textwidth}{@{}>{\raggedright\arraybackslash}p{.16\textwidth}Y>{\raggedright\arraybackslash}p{.21\textwidth}@{}}
\toprule
Task / source & Measurement map and target & Evaluation \\
\midrule
Panda Cartesian reach (R0) & Position $p_{\rm EE}(T)/L_P$; orientation error $\rho_T/L_R$, $\rho_T=\operatorname{rotvec}(R_T,R_0)$; terminal $v_{\rm EE}/L_V$ and $\omega_{\rm EE}/L_W$. \emph{Target:} Move position; preserve orientation and terminal linear/angular velocity & Executed Cartesian target error \\
Panda checkpoint/via-point reach (R1) & $[p_{\rm EE}(t_1),\ldots,p_{\rm EE}(T)]/L_P$, with $Z=3,6,$ or $9$. \emph{Target:} Independent checkpoint displacements or geometry-defined via-points & Executed closest-approach via error and final error \\
Reach with orientation hold (R0/R1) & $[p_{\rm EE}(T)/L_P,\operatorname{rotvec}(R_T,R_0)/L_R]$. \emph{Target:} Move position; request zero orientation change & Executed target error and tool rotation \\
Lateral move at constant height (R0/R1) & Terminal $(x_{\rm EE},y_{\rm EE})/L_P$ and four checkpoint heights $z_{\rm EE}(t_i)/L_P$. \emph{Target:} Move laterally; hold start height at four times & Executed target error and maximum height drift \\
Unmodelled obstacle reach (R0) & $[p_{\rm EE}(T)/L_P,\operatorname{rotvec}/L_R,c(\tau)/L_C]$ with soft-minimum whole-arm clearance. \emph{Target:} Request goal and clearance; preserve orientation & Contact-free execution and MuJoCo geom distance \\
Spatial locomotion target (R0/R1) & Midpoint/final $x$, terminal $v_x$, torso height, and pitch, each with its frozen physical scale. \emph{Target:} Request intermediate/final position and speed; preserve height and pitch & Executed world-position error \\
Locomotion trajectory tracking (R0/R1) & Four $x(t_i)$ checkpoints, terminal $v_x$, torso height, and pitch, each frozen-scaled. \emph{Target:} Request a smooth world-position profile; preserve posture & Executed trajectory RMSE \\
Progress and speed control (R0) & Progress $\Delta x/L_X$, terminal speed $v_x/L_V$, height $z/L_Z$, and pitch$/L_{\rm pitch}$. \emph{Target:} Request signed progress/speed change; preserve height and pitch & Executed progress or signed speed change \\
\bottomrule
\end{tabularx}
\end{table*}

\subsection{Control and planning}
The catalogue first defines character motion generation requests, then the
canonical control and trajectory-planning objectives. Both combine requested
motion with explicitly represented preservation or terminal-state requirements.

\subsubsection{Character motion generation}\label{sup:sec:character-objectives}
Character steering differentiates joint coordinates recovered by forward kinematics
through the remaining motion sampler. Obstacle avoidance specifies a mid-clip pelvis
shift around a 20 cm sphere, preserving final pelvis location and seven arm-bone lengths
($Z=13$). Reaching specifies five wrist waypoints, preserved ankle and hip positions at two additional frames,
zero wrist arrival velocity, and bone lengths at waypoint frames ($Z=77$).
Requested displacements use the unsteered subject frame and natural excursion scales.
The reported conjunctive success criterion requires positional target accuracy,
preservation of joint positions, and jerk below a population-calibrated threshold.
Velocity, bone drift, and geometric collision are separate diagnostics, even when
the corresponding quantities enter the steering objective. The full campaign definitions and reported thresholds are in
Appendix~\ref{app:motion}.

\subsubsection{Canonical control and trajectory planning}

Table~\ref{sup:tab:control-planning-objectives} lists the canonical control and planning objectives. Here $h$ is evaluated on a generated state trajectory.  Endpoint terms encode the requested final
state; checkpoint terms encode the route; and velocity, posture, effort, and clearance
entries specify either absolute targets or changes relative to the anchor.  Collision and execution diagnostics are evaluated according to each task protocol, so satisfying a sparse endpoint request alone is not reported as obstacle-avoidance success.

\begin{table*}[!htbp]
\centering\small
\caption{Control and planning objective and cost functions.  State coordinates are normalized by the frozen physical scales named in the source protocols before entering the Euclidean request cost.}
\label{sup:tab:control-planning-objectives}
\setlength{\tabcolsep}{3pt}
\begin{tabularx}{\textwidth}{@{}>{\raggedright\arraybackslash}p{.16\textwidth}Y>{\raggedright\arraybackslash}p{.21\textwidth}@{}}
\toprule
Task / source & Measurement map and target & Evaluation \\
\midrule
Pendulum stabilization (C0) & $[\sin(\theta_T-\theta^\star),\ 1-\cos(\theta_T-\theta^\star),\ \dot\theta_T,\ E_u]$. \emph{Target:} Target $[0,0,0,E_{u,0}]$; control effort is preserved & Real-dynamics replay: angle, speed, return, and effort \\
Pendulum reference tracking (C0) & Wrapped angle errors at fixed intermediate checkpoints and $T$, plus $\dot\theta_T$. \emph{Target:} Reach the declared reference angles and finish stationary & Real-dynamics replay and tracking error \\
CartPole target position (C0) & $[x_T,\theta_T,\dot x_T,\dot\theta_T]$ divided by frozen training-distribution scales. \emph{Target:} Target $[x^\star,0,0,0]$; upright orientation and terminal velocities are explicit zeros & Open-loop true-dynamics survival and terminal errors \\
Double-integrator waypoints (P0) & $[p_{t_1},\ldots,p_{t_K},p_T,v_T]$, $K\in\{1,2,4,8\}$. \emph{Target:} Hit all requested positions; finish at $v_T=0$ & Executed goal/waypoint error and task success \\
Maze2D goal and via-points (P0) & $[p_{t_1},\ldots,p_{t_K},p_T,v_T]$ with A* reference waypoints. \emph{Target:} Reach requested free-space waypoints and goal; finish stationary & Executed goal distance, wall collision, path length \\
Dense obstacle navigation (P0) & $[p_T,v_T,\operatorname{viol}_{1:H},s]$, $\operatorname{viol}_t=\sum_j[r_j+r_{\rm safe}-\|p_t-o_j\|]_+$. \emph{Target:} Request the new goal and zero velocity/clearance violations; preserve smoothness $s$ & Executed collision, clearance, goal error \\
Planar-arm pose planning (P0) & $[p_{\rm EE,T},\cos\phi_T,\sin\phi_T,\dot p_{\rm EE,T}]$; dense body-clearance entries are added for the obstacle task. \emph{Target:} Move the end effector; preserve orientation, terminal velocity, and clearance & Executed pose error and body collision \\
\bottomrule
\end{tabularx}
\end{table*}

\end{document}